\documentclass{article}

\usepackage{microtype}
\usepackage{graphicx}
\usepackage{caption}
\usepackage{subcaption}
\usepackage{booktabs} % for professional tables
\usepackage{makecell}
\usepackage{multirow}
\usepackage{tabularx}
\usepackage{multicol}

\usepackage{xcolor}

\usepackage{listings}

\definecolor{dkred}{rgb}{0.5,0,0}
\definecolor{dkgreen}{rgb}{0,0.6,0}
\definecolor{gray}{rgb}{0.5,0.5,0.5}
\definecolor{mauve}{rgb}{0.58,0,0.82}
\newcommand{\system}{{\sc SUOD}\xspace}

\usepackage{hyperref}

\usepackage[utf8]{inputenc}
\usepackage{amsmath}
\usepackage{physics}
\usepackage{amsopn}

\usepackage{amsfonts}
\usepackage[absolute]{textpos}
\usepackage{bm}
\usepackage{xspace}
\usepackage{bbm}
\usepackage{perpage}
\MakePerPage{footnote}
\usepackage{paralist}

\usepackage[newfloat,frozencache,cachedir=.]{minted}

\usepackage[accepted]{mlsys2021}

\mlsystitlerunning{\system: Accelerating Large-Scale Unsupervised Heterogeneous Outlier Detection}

\begin{document}

\twocolumn[
\mlsystitle{\system: Accelerating Large-Scale Unsupervised Heterogeneous Outlier Detection}

% unless you've provided the [accepted] option to the mlsys2020


\mlsyssetsymbol{equal}{*}

\begin{mlsysauthorlist}
\mlsysauthor{Yue Zhao}{equal,cmu}
\mlsysauthor{Xiyang Hu}{equal,cmu}
\mlsysauthor{Cheng Cheng}{cmu}
\mlsysauthor{Cong Wang}{pku}
\mlsysauthor{Changlin Wan}{purdue}
\mlsysauthor{Wen Wang}{cmu}
\mlsysauthor{Jianing Yang}{cmu}
\mlsysauthor{Haoping Bai}{cmu}
\mlsysauthor{Zheng Li}{arima}
\mlsysauthor{Cao Xiao}{iqvia}
\mlsysauthor{Yunlong Wang}{iqvia}
\mlsysauthor{Zhi Qiao}{iqvia}
\mlsysauthor{Jimeng Sun}{uiuc}
\mlsysauthor{Leman Akoglu}{cmu}
\end{mlsysauthorlist}

\mlsysaffiliation{cmu}{Carnegie Mellon University}
\mlsysaffiliation{iqvia}{IQVIA}
\mlsysaffiliation{pku}{Peking University}
\mlsysaffiliation{purdue}{Purdue University}
\mlsysaffiliation{uiuc}{University of Illinois at Urbana-Champaign}
\mlsysaffiliation{arima}{Arima Inc.}

\mlsyscorrespondingauthor{Yue Zhao}{zhaoy@cmu.edu}
\mlsyscorrespondingauthor{Xiyang Hu}{xiyanghu@cmu.edu}

\mlsyskeywords{Machine Learning, MLSys, Outlier Detection, Anomaly Detection}

\vskip 0.3in

\begin{abstract}

Outlier detection (OD) is a key machine learning (ML) task for identifying abnormal objects from general samples with numerous high-stake applications including fraud detection and intrusion detection. Due to the lack of ground truth labels, practitioners often have to build a large number of unsupervised, heterogeneous models (i.e., different algorithms with varying hyperparameters) for further combination and analysis, rather than relying on a single model. How to \textit{accelerate} the training and scoring on new-coming samples by outlyingness (referred as prediction throughout the paper) with \textit{a large number of unsupervised, heterogeneous} OD models? 
In this study, we propose a modular acceleration system, called \system, to address it. The proposed system focuses on three complementary acceleration aspects (data reduction for high-dimensional data, approximation for costly models, and taskload imbalance optimization for distributed environment), while maintaining performance accuracy.  
Extensive experiments on more than 20 benchmark datasets demonstrate \system's effectiveness in heterogeneous OD acceleration, along with a real-world deployment case on fraudulent claim analysis at IQVIA, a leading healthcare firm.
We open-source \system for reproducibility and accessibility.
\end{abstract}
]



\printAffiliationsAndNotice{\mlsysEqualContribution} % otherwise use the standard text.

\section{Introduction}
Outlier detection (OD) aims at identifying the samples that are deviant from the general data distribution \cite{zhao2019pyod,DBLP:journals/corr/abs-2009-09822}, which has been used in various applications \cite{Chandola2009Anomaly,DBLP:conf/icdm/ZhaLWH20}. Notably, most of the existing OD algorithms are unsupervised due to the high cost of acquiring ground truth \cite{zhao2019lscp}. Model selection and hyperparameter tuning in OD
have been shown to be non-trivial problems
\cite{zhao2020automating,DBLP:conf/icdm/LiZBIH20}. To reduce the risk and instability of using a single OD model, practitioners prefer to build a large corpus of OD models with variation and diversity, e.g., different algorithms, varying parameters, distinct views of the datasets, etc \cite{Aggarwal2017}. This approach is known as \textit{heterogeneous OD}. Ensemble methods that select and combine diversified base models can be leveraged to analyze heterogeneous OD models \cite{Aggarwal2013,zimek2014ensembles,Aggarwal2017}, and more reliable results may be achieved.
The simplest combination is to take the average or maximum across all the base models as the final result \cite{Aggarwal2017}, along with more complex combination approaches in both unsupervised \cite{zhao2019lscp} and semi-supervised manners \cite{zhao2018xgbod}. 

However, training and prediction with a large number of heterogeneous OD models is computationally expensive on high-dimensional, large datasets. For instance, proximity-based algorithms, assuming outliers behave differently in specific regions \cite{aggarwal2016outlier}, can be prohibitively slow or even completely fail to work under this setting.
Representative methods such as $k$ nearest neighbors ($k$NN) \cite{Ramaswamy2000}, local outlier factor (LOF) \cite{Breunig2000}, and local outlier probabilities (LoOP) \cite{Kriegel2009}, operate in Euclidean space for distance/density calculation, suffering from the curse of dimensionality \cite{schubert2015fast}. Numerous works have attempted to tackle this scalability challenge from various angles, e.g., data projection \cite{keller2012hics}, subspacing \cite{liu2008isolation}, and distributed learning for specific OD algorithms \cite{lozano2005parallel,oku2014parallel}. \textbf{However, none of them provides a comprehensive solution by considering all aspects of large-scale heterogeneous OD, leading to limited practicability and efficacy}.

\begin{figure}[!htp]
\centering
    \includegraphics[width=\linewidth]{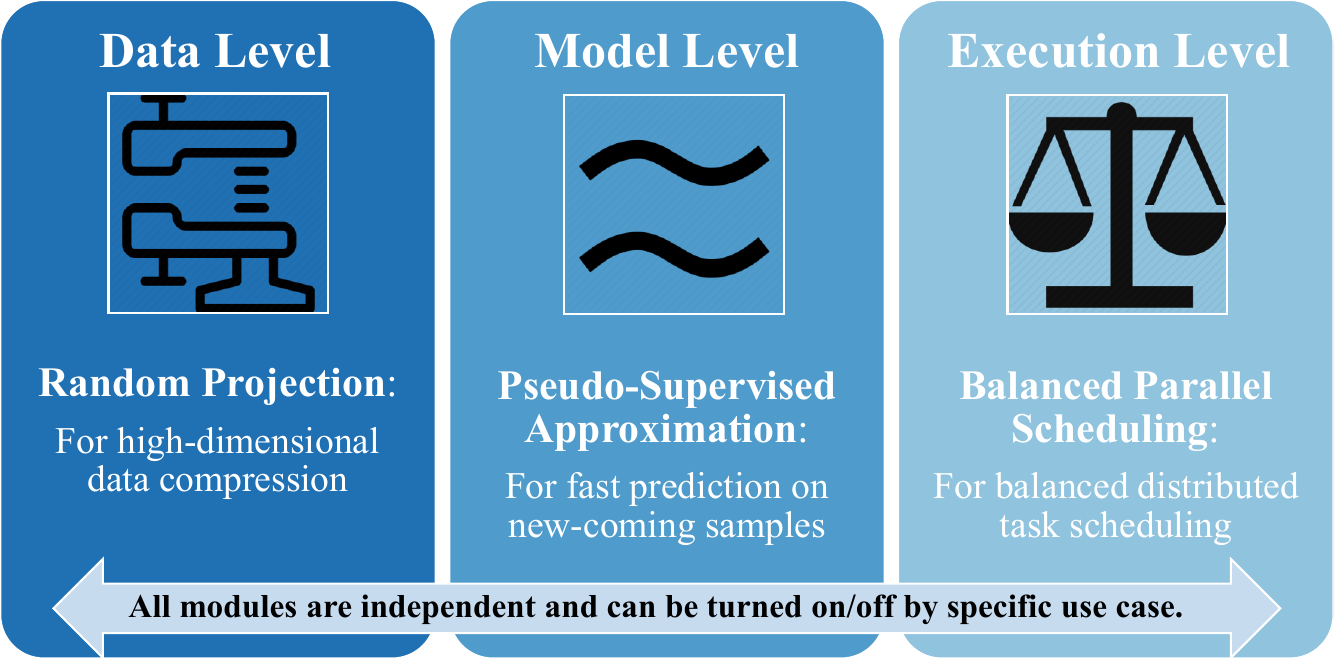}
\caption{\system focuses on three independent levels.} 
\label{fig:system}
\end{figure}

To tap the gap, we propose a comprehensive acceleration framework called \system.
As shown in Fig. \ref{fig:system}, \system has three modules focusing on complementary levels: random projection (\textbf{data level}), pseudo-supervised approximation (\textbf{model level}), and balanced parallel scheduling (\textbf{execution level}). For high-dimensional data, \system generates a random low-dimensional subspace for each base model by Johnson-Lindenstrauss projection \cite{johnson1984extensions}, in which the corresponding base model is trained. If prediction on new-coming samples is needed, fast supervised models are employed to approximate costly unsupervised outlier detectors (e.g., \textit{k}NN and LOF). 
To train the supervised approximators, we regard the unsupervised models' outputs on the train set as ``pseudo ground truth". Intuitively, this may be viewed as distilling knowledge from complex unsupervised models \cite{hinton2015distilling} by fast and more interpretable supervised models.  
We also build a taskload predictor to reduce the scheduling imbalance in distributed environment. Other than generically assigning the equal number of models to each worker, our balanced parallel scheduling mechanism forecasts OD model cost, e.g., training time, before scheduling them, so that the taskload is evenly distributed among workers. 
Notably, all three acceleration modules are designed to be independent but complementary, which can be used alone or combined as a system. It is noted that \system is designed for offline learning with a \textit{stationary assumption}, although it may be further extended to \textit{online setting} for streaming data with extra effort \cite{wang2019robust,wang2020few}. It is noted that offline training and prediction is one of the primary scenarios in machine learning applications \cite{Amazon,DBLP:conf/osdi/NakandalaSYKCWI20}. 

Our contributions are summarized as follows:
\begin{compactenum}
    \item \textbf{The First Comprehensive OD Acceleration System}: We propose \system, (to our knowledge) the most comprehensive system for heterogeneous OD acceleration by a holistic view on data, model, and execution level.
    \item \textbf{Analysis of Data Compression}: We examine various data compression methods and identify the %performing 
    best performing
    method(s) for large-scale outlier ensembles.
    \item \textbf{Exploration of Model Approximation for OD}: We analyze the use of pseudo-supervised regression models' performance in approximating costly unsupervised OD models, as the first research effort on this topic.
    % To our best knowledge, this is the first research effort on this topic.
    \item \textbf{Forecasting-based Scheduling System}: We fix an imbalance scheduling issue in distributed heterogeneous OD efficiency, saving up to 61\% execution time.
    \item \textbf{Effectiveness}: We conduct extensive experiments to show the effectiveness of the acceleration modules independently, and of the entire framework as a whole, along with a real-world case on fraud detection.
    % \item \textbf{Open-source System}: We release \system with industry-level implementation for accessibility and reproducibility\footnote{\url{https://github.com/yzhao062/SUOD}}, also as a core component of 
    % % the leading OD library 
    % PyOD\footnote{\url{https://github.com/yzhao062/PyOD}}.
    % By the submission time, it has been downloaded by more than 700,000 times with real-world deployment cases\footnote{\url{https://github.com/yzhao062/SUOD}}. 
    % It is also integrated as a core component of the leading OD library PyOD\footnote{\url{https://github.com/yzhao062/PyOD}}.
    % \item \yz{Discuss how to customize each module to tailor different needs, and the possibility of extending the framework to general unsupervised learning tasks, beyond the scope of outlier detection.}
    % \item To foster reproducibility and accessibility, the framework, all code, figures, and datasets are openly shared\footnote{\url{https://github.com/yzhao062/SUOD}}. 
\end{compactenum}

To foster accessibility and reproducibility, We release \system with industry-level implementation\footnote{\url{https://github.com/yzhao062/SUOD}}, which also becomes a core component of the leading OD library PyOD\footnote{\url{https://github.com/yzhao062/PyOD}}.

\section{Related Work}

\subsection{Outlier Detection and Ensemble Learning}
Outlier detection has numerous important applications, such as rare disease detection \cite{li2018semi}, healthcare utilization analysis \cite{hu2012healthcare}, video surveillance \cite{lu2017unsupervised}, fraudulent online review analysis \cite{akoglu2013opinion}, and network intrusion detection \cite{lazarevic2003comparative}. Yet, detecting outliers is challenging due to various reasons \cite{Aggarwal2013,zhao2019lscp,zhao2019pyod}. 
Most of the existing detection algorithms are unsupervised as ground truth is often absent in practice, and acquiring labels can be prohibitively expensive.  We include 8 popular OD algorithms in this study for experimentation, including Isolation Forest \cite{liu2008isolation}, Local Outlier Factor (LOF) \cite{Breunig2000}, Angle-based Outlier Detection (ABOD) \cite{Kriegel2009}, Feature Bagging \cite{lazarevic2005feature}, Histogram-based Outlier Score (HBOS) \cite{goldstein2012histogram}, and Clustering-Based Local Outlier Factor (CBLOF) \cite{he2003discovering}. See Appendix \ref{appendix:model_set} for details on OD models.

Consequently, relying on a single unsupervised model has inherently high risk, and outlier ensembles that leverage a group of diversified (e.g., heterogeneous) detectors have become increasingly popular \cite{Aggarwal2013,zimek2014ensembles,Aggarwal2017}. There are a group of unsupervised outlier ensemble frameworks proposed in the last several years from simple combination like averaging \cite{Aggarwal2017} to more complex model selection approaches like SELECT \cite{rayana2016less}, LSCP \cite{zhao2019lscp}, and MetaOD \cite{zhao2020automating}. 
In addition to fully unsupervised outlier ensembles, there are (semi-)supervised ensembling frameworks that can incorporate existing label information such like XGBOD \cite{zhao2018xgbod}.
For both unsupervised and (semi-)supervised methods, a large group of unsupervised, heterogeneous OD models are used as base for robustness and performance---\system is hereby proposed to accelerate for this scenario.

\subsection{Scalability and Efficiency Challenges in OD}
\label{subsec:scalability_issues}
Efforts have been made through various channels to accelerate large-scale OD. On the \textbf{data level}, researchers try to project high-dimensional data onto lower-dimensional subspaces \cite{achlioptas2001database}, including simple Principal Component Analysis (PCA) \cite{shyu2003novel} and more complex subspace method HiCS \cite{keller2012hics}. However, deterministic projection methods, e.g., PCA, are not ideal for building diversified heterogeneous OD---they lead to the same or similar subspaces with limited diversity by nature, resulting in the loss of outliers \cite{aggarwal2016outlier}.
Complex projection and subspace methods may bring performance improvement for outlier mining, but the generalization capacity is limited. Hence, projection methods preserving pairwise distance relationships for downstream tasks should be considered. \system's considers both diversity induction and pairwise distance preservation for downstream tasks, leading to diversified and meaningful feature spaces (see \S \ref{sec:RP}).

On the \textbf{model level}, knowledge distillation emerges as a good way of compressing large neural networks \cite{hinton2015distilling}, while its usage in outlier detection is still underexplored. Knowledge distillation refers to the notion of compressing a large, often cumbersome model(s) into a small and more interpretable one(s). 
Under the context of OD, proximity-based models, such as LOF, can be slow (high time complexity) for predicting on new-coming samples with limited interpretability, which severely restricts their usability. \system adapts a similar idea to outlier mining by ``distilling" complex unsupervised models. Although \system shares a similar concept as knowledge distillation for computational cost optimization, there are a few notable differences (see \S \ref{sec:psa}).

There are also engineering cures on the \textbf{execution level}. For various reasons, OD has no mature and efficient distributed frameworks with thousands of clusters---distributed computing for OD mainly falls into the category of ``scale-up" that focuses on leveraging multiple local cores on a single machine more efficiently.
To this end, specific OD algorithms can be accelerated by distributed computing with multiple workers (e.g., CPU cores) \cite{oku2014parallel,lozano2005parallel}. However, these frameworks are not designed for a group of heterogeneous models but only a single algorithm, which limits their usability. 
It is noted that a group of heterogeneous detection models can have significantly varied computational cost. As a simple example, let us split 100 heterogeneous models into 4 groups for parallel training. If group \#2 takes significantly longer time than the others to finish, it behaves like the bottleneck of the system. More formally, imbalanced task scheduling causes the system efficiency to be curbed by the worker taking the most time.
There are also a line of system researches focus on more efficient task scheduling for ``shorter tasks on larger degree of parallelism", e.g., Sparrow \cite{DBLP:conf/sosp/OusterhoutWZS13} and Pigeon \cite{DBLP:conf/cloud/0001LLSRCJ19}. For instance, Sparrow discusses scheduling millions of tasks (at millisecond scale) on thousands of machines.
Differently, Heterogeneous OD applications typically use a few OD models (e.g., 5 to 1,000) and each of them takes a few seconds to hours
to run (a considerable number of tasks; each takes time to run), which is ``longer tasks with a small number of workers". 
\system is, therefore, proposed to reduce the inefficiency in distributed heterogeneous OD specifically.

\section{System Design}
\label{Sec:Algorithm_Design}

\subsection{Problem Formulation}
OD applications and research are primarily running a single, on-prime, powerful machine with multiple cores/workers, due to its high-stake nature (e.g., data sensitive of financial transactions).
Therefore, we formulate unsupervised heterogeneous OD training and prediction tasks with:
\begin{compactitem}
\item $m$ unsupervised heterogeneous OD models, where  $\mathcal{M}=\{M_1,...,M_m\}$.
We refer the combination of an algorithm and its  hyperparamters as a model.
\item train data $\mathbf{X}_{\text{train}} \in \mathbbm{R}^{n \times d}$ without ground truth labels.
\item (optional) test data  $\mathbf{X}_{\text{test}} \in \mathbbm{R}^{l \times d}$ for prediction. The OD models should be trained first.
\item (optional) $t$ available workers for distributed computing, e.g., $t$ cores on a single machine. This constructs the worker pool as $\mathcal{W}=\{W_1,...,W_t\}$. By default, a single worker setting ($t=1$) is assumed. 
\end{compactitem}

Without \system, one will train each model in $\mathcal{M}$ on $\mathbf{X}_{\text{train}}$ iteratively, e.g., with a for loop. If there are multiple workers available ($t>1$), a generic scheduling system will equally split $m$ models into $t$ groups, so each available worker will process roughly $\lceil \frac{m}{t} \rceil $ models. Prediction on new-coming samples $\mathbf{X}_{\text{test}}$ follows the similar manner as training. 
See Algorithm \ref{alg:SUOD} for detailed symbol definition.

\renewcommand{\algorithmicrequire}{\textbf{Input:}}
\renewcommand{\algorithmicensure}{\textbf{Output:}}
\renewcommand{\algorithmiccomment}[1]{\hfill$\blacktriangleright$ #1}
\newcommand{\cdash}{\multicolumn{1}{c}{--} }

\begin{algorithm} [!ht]
	\caption{\system: Training and Prediction}
	\label{alg:SUOD}
	\begin{algorithmic}[1]
	    \small{
		\REQUIRE $m$ unsupervised OD models $\mathcal{M}$; train data $\mathbf{X}_{\text{train}} \in \mathbbm{R}^{n \times d}$; target dimension $k$; the number of available workers $t$ (optional, default to 1); test data  $\mathbf{X}_{\text{test}} \in \mathbbm{R}^{l \times d}$ (optional); supervised regressor $R$ (optional)
		\ENSURE Trained OD models $\mathcal{M}$; fitted pseudo-supervised regressors $R$ (optional); test prediction results $\hat{y}_{\text{test}}$ (optional)
		\\\hrulefill
		\FOR{each model $M_i$ in $\mathcal{M}$}
		\IF{random projection is enabled (\S \ref{sec:RP})} 
		\STATE Initialize a JL transformation matrix $\mathbf{W} \in \mathbbm{R}^{d\times k}$
		\STATE Get feature subspace $\psi_{i}  \mathrel{\mathop:}= \langle \mathbf{X}_{\text{train}}, \mathbf{W} \rangle \in \mathbbm{R}^{n \times k}$
		\ELSE%{ $d <\theta$} \COMMENT{disable random projection}
		\STATE Use the original space $\psi_{i} \mathrel{\mathop:}=\mathbf{X}_{\text{train}} \in \mathbbm{R}^{n \times d}$
		\ENDIF
		\ENDFOR
		\IF{number of available workers $t > 1$}  %\COMMENT{parallel learning activated}
		\STATE Scheduling the training of $m$ models to $t$ workers by minimizing Eq. \ref{eq:balanced_learning_goal} (see \S \ref{sec:bps}). Models are trained on the corresponding feature space $[\psi_{1}, ..., \psi_{m}]$.
		\ELSE 
		\STATE Train each model $M_i$ in $\mathcal{M}$ on its corresponding $\psi_{i}$.
		\ENDIF
		\STATE Return trained models $\mathcal{M}$
		\\\hrulefill
		\IF{Scoring on newcoming samples $\mathbf{X}_{\text{test}}$}     %\COMMENT{model approximation activated}
		\STATE Acquire the pseudo ground truth $target^{\psi_{i}}$ as the output of $M_i$ on $\psi_{i}$, i.e., $target^{\psi_{i}} \mathrel{\mathop:}= M_i(\psi_{i})$ 
		\FOR{each costly model $M_i$ in $\mathcal{M}_c$}
% 		\STATE \COMMENT{Expedite loop by Balanced Parallel Scheduling (\S \ref{sec:bps})}
        \STATE Initialize a supervised regressor $R_i$
		\STATE Train $R_i$ by $\{\psi_{i}, target^{\psi_{i}}\}$ (see \S \ref{sec:psa})
		\STATE Predict by supervised $R_i$,  $\hat{y}_{\text{test}}^{i}=R_i.\textrm{predict}(\mathbf{X}_{\text{test}})$ 
% 		\COMMENT{scoring by approximators}
		\ENDFOR
		\STATE Return $\hat{y}_{\text{test}}$ and approximation regressors $R$
		\ENDIF}
	\end{algorithmic}
\end{algorithm}

\subsection{System Design}
\system is designed to accelerate the above procedures with three independent modules targeting different levels (data, model, and execution). \textbf{Each module can be flexibly enabled or disabled} as shown in Algorithm \ref{alg:SUOD}. 
For high-dimensional data, \system can randomly project the original feature onto low-dimensional spaces (\S \ref{sec:RP}). Pairwise distance relationships are expected to be maintained, and the diversity is induced for ensemble construction. A fast supervised regressor could be used to approximate the output of each slow and costly unsupervised detector. We could use the supervised regressor for fast prediction (\S \ref{sec:psa}). If there are multiple available workers for distributed computing, we propose a forecasting-based scheduling mechanism (\S \ref{sec:bps}) to reduce taskload imbalance in heterogeneous OD.
% expedite the training and prediction execution with a large number of heterogeneous models.

\system's API design follows \texttt{scikit-learn} \cite{pedregosa2011scikit} and \texttt{PyOD} \cite{zhao2019pyod}, with an \textit{initialization}, \textit{fit}, and \textit{prediction} schema (see Codeblock \ref{code:api}).

\newenvironment{code}{\captionsetup{type=listing}}{}
\SetupFloatingEnvironment{listing}{name=Codeblock}

\begin{code}
\begin{minted}[frame=single,
% framesep=2mm,
% baselinestretch=1,
fontsize=\scriptsize % footnotesize
]{python}
import SUOD
# initialize a group of heterogeneous OD models
base_estimators=[
    LOF(n_neighbors=40), ABOD(n_neighbors=50), 
    LOF(n_neighbors=60), IForest(n_estimators=100)]  
    
# initialize SUOD with module flags
clf=SUOD(base_estimators=base_estimators, 
    rp_flag_global=True, 
    approx_clf=RandomForestRegressor(), 
    bps_flag=True, 
    approx_flag_global=True)
    
# fit and make prediction
clf.fit(X_train)                                                             
test_labels=clf.predict(X_test)  
test_scores=clf.decision_function(X_test)}
\end{minted}
\captionof{listing}{\system's easy-to-use API (inspired by scikit-learn)}
\label{code:api}
\end{code}

\subsection{Data Level: Random Projection (RP) for Data Compression} 
\label{sec:RP}
For high-dimensional datasets, many proximity-based OD algorithms suffer from the curse of dimensionality \cite{lazarevic2005feature}. A widely used dimensionality reduction to cure this is the Johnson-Lindenstrauss (JL) projection \cite{johnson1984extensions}, which has been applied to OD because of its great scalability \cite{schubert2015fast}. Unlike PCA discussed in \S \ref{subsec:scalability_issues}, JL projection could compress the data without heavy distortion on the Euclidean space---\textit{outlyingness information is therefore preserved in the compression}. Moreover, its built-in randomness can be useful for diversity induction in heterogeneous OD---data randomness can also serve as a source of heterogeneity. Additionally, JL projection ($\mathcal{O} (ndk)$) is more efficient than popular PCA ($\mathcal{O} (nd^2+n^3)$) with lower time complexity, where $k$ is the target dimension for compression.

JL projection is defined as: given a set of $n$ samples  $\mathbf{X} = \{\mathbf{x}_1, \mathbf{x}_2, ... \mathbf{x}_n\},$ each $\mathbf{x}_i \in \mathbbm{R}^d$, let $\mathbf{W}$ be a $k \times d$ projection matrix with each entry drawing independently from a predefined distribution $F$, e.g., Gaussian, so that $\mathbf{W} \sim F$. Then the JL projection is a function $f: \mathbbm{R}^d \rightarrow \mathbbm{R}^k$ such that
$f(\mathbf{x}_i) = \frac{1}{\sqrt{k}}\mathbf{x}_i\mathbf{W}^T$. JL projection randomly projects high-dimensional data ($d$ dimensions) to lower-dimensional subspaces ($k$ dimensions), but preserves the distance relationship between points. 
In fact, if we fix some $\mathbf{v} \in \mathbbm{R}^d$, 
% and let $\mathbf{W}$ be the $k \times d$ matrix such that each entry is from $\mathcal{N}(0,1)$. 
for every $\epsilon \in (0,3)$, we have \cite{schubert2015fast}:

\begin{equation}
\label{eq_JL}
% \scriptsize
\footnotesize
% \tiny
P\left[(1-\epsilon){\Vert \mathbf{v} \Vert}^2 \leq {\Vert \frac{1}{\sqrt{k}}\mathbf{v}\mathbf{W}^T \Vert}^2 \leq (1 + \epsilon) {\Vert \mathbf{v} \Vert}^2 \right] \leq 2e^{-\epsilon^2 \frac{k}{6}}
\end{equation} 

Let $\mathbf{v}$ to be the differences between vectors. Then, the above bound shows that for a finite set of $n$ vectors $\mathbf{X} = \{\mathbf{x}_1, \mathbf{x}_2, ... \mathbf{x}_n\} \in \mathbbm{R}^d$, the pairwise Euclidean distance is preserved within a factor of $(1 \pm \epsilon)$, reducing the vectors to $k = \mathcal{O}(\frac{log(n)}{\epsilon ^2})$ dimensions.

Four distributions $F$ for JL projection are considered in this study: (i) \textit{basic}: the transformation matrix is generated by standard Gaussian; (ii) \textit{discrete}: the transformation matrix is picked randomly from Rademacher distribution (uniform in $\{-1,1\})$; (iii) \textit{circulant}: the transformation matrix is obtained by rotating the subsequent rows from the first row which is generated from standard Gaussian and (iv) \textit{toeplitz}: the first row and column of the transformation matrix are generated from standard Gaussian, and each diagonal uses a constant value from the first row and column. A more thorough empirical study on JL methods can be found in \cite{venkatasubramanian2011johnson}.

For $\mathbf{X}_{\text{train}}$, RP can reduce the original feature space $d$ to the target dimension $k$. Specifically, \system initializes a JL transformation matrix $\mathbf{W} \in \mathbbm{R}^{d\times k}$ by drawing from one of the four distributions $F$. $\mathbf{X}_{\text{train}}$ is therefore projected onto the $k$ dimension feature space as  $\mathbf{X}_{\text{train}}' = \langle \mathbf{X}_{\text{train}}, \mathbf{W} \rangle \in \mathbbm{R}^{n \times k}$. The transformation matrix $\mathbf{W}$ should be kept for transforming newcoming samples: $\mathbf{X}_{\text{test}}' = \langle \mathbf{X}_{\text{test}}, \mathbf{W} \rangle \in \mathbbm{R}^{m \times k}$. It is noted that RP module should be used with caution. First, projection may not be helpful or even detrimental for subspace methods like Isolation Forest and HBOS. Second, if the number of samples $n$ is too small, the bound in Eq. (\ref{eq_JL}) does not hold.
% ---the result may be less stable.

\subsection{Model Level: Pseudo-Supervised Approximation (PSA) for Fast Prediction}
\label{sec:psa}
PSA module is designed to speed up prediction on new-coming samples.
Specifically, after the models in $\mathcal{M}$ are trained, 
\system uses PSA to approximate and replace each \textbf{costly unsupervised model} by a \textbf{faster supervised regressor} for \textbf{fast offline prediction}. Notably, only costly unsupervised models should be replaced; the cost can be measured through time complexity analysis.
For instance, proximity-based algorithms like $k$NN and LOF are costly in prediction  (upper bounded by $\mathcal{O} (nd)$), and can be effectively replaced by ``cheaper" supervised models like random forest \cite{breiman2001random} (upper bounded by $\mathcal{O} (ph)$ where $p$ denotes the number of base trees and $h$ denotes the max depth of a tree; often $p \ll n$ and $h \leq d$). This ``pseudo-supervised" model uses the output of unsupervised models (outlyingness score) as ``the pseudo ground truth"---\textit{the goal is to approximate the output of the underlying unsupervised model}. It is noted that the approximator's prediction cost (i.e., time complexity) must be lower than the underlying unsupervised model, while maintaining a comparable level of prediction accuracy. For instance, fast (low time complexity) OD algorithms like Isolation Forest and HBOS should not be approximated and replaced. To facilitate this process, we predefine the pool of costly OD algorithm $\mathcal{M}_c$. If a model $M_i$ is in  $\mathcal{M}_c$, it will be approximated by default.

As shown in Algorithm \ref{alg:SUOD}, for each
costly trained unsupervised model $M_i$ belonging to $\mathcal{M}_c$, a supervised regressor $R_i$ is trained by $\{\mathbf{X}_{\text{train}}, \mathbf{y}_i \}$; $\mathbf{y}_i$ is the outlyingness score by $M_i$ on the train set (referred as pseudo ground truth)\footnote{If RP is enabled, $\mathbf{X}_{\text{train}}$ is replaced by the compressed space.}. $R_i$ is then used to predict on unseen data $\mathbf{X}_{\text{test}}$. 

\textbf{Remark 1:} Supervised tree ensembles are recommended for PSA due to their outstanding scalability, robustness to overfitting, and interpretability (e.g., feature importance) \cite{hu2019optimal}. In addition to the execution time reduction, supervised models are generally more interpretable. For instance, random forest used in the experiments can yield feature importance automatically to facilitate understanding. 
% \la{I recommend making a list of all heterogeneous detectors that SUOD employs as rows in a Table and report  train and test complexity in 2 separate columns in big(O) notation. Then you can even split them into costly vs. fast (and make clear you refer to test/scoring time).} 

\textbf{Remark 2:} Notably, PSA may be viewed as using supervised regressors to distill knowledge from unsupervised OD models. 
% However, it is different from the established knowledge distillation in multiple aspects. First, 
However, it works in a fully unsupervised manner, unlike the classic distillation under supervised settings. 

\begin{figure}[t]
\centering
    \includegraphics[width=\linewidth]{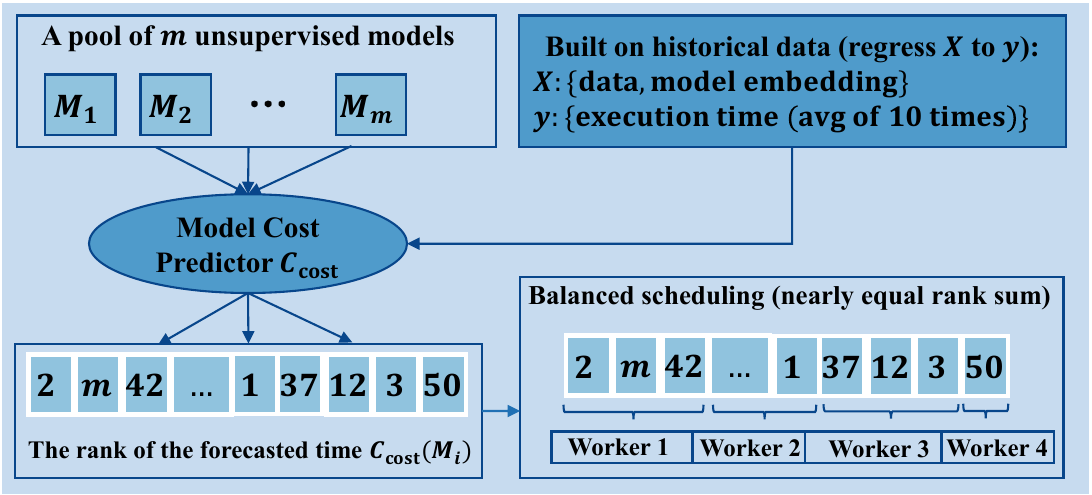}
\caption{\label{fig:flowchart}Flowchart of balanced parallel scheduling, which aims to assign nearly equal rank sum by model cost predictor $C_{\text{cost}}$.} 
\label{fig:balanceLearning}
\end{figure}  

\subsection{Execution level: Balanced Parallel Scheduling (BPS) for Taskload Imbalance Reduction}
\label{sec:bps}

% \subsubsection{Taskload Imbalance within Distributed Systems}
\textbf{Taskload Imbalance within Distributed Systems}: It is likely to observe 
system inefficiency due to taskload imbalance among workers when a large number of heterogeneous models are used.
% to train or predict with a large number of heterogeneous models in a distributed environment, practitioners may use generic and simple scheduling by scheduling equal number of models to each worker, which can cause severe system inefficiency due to taskload imbalance.
% \textit{Without BPS, practitioners often use generic scheduling by distributing equal number of models to each worker.} 
% It is noted taskload imbalance among workers curbs an efficient execution.
For instance, we want to train 25 OD models with varying parameters from each of the four algorithms \{$k$NN, Isolation Forest, HBOS, OCSVM\} (100 heterogeneous models in total).
The existing distributed frameworks, e.g., the voting machine in scikit-learn \cite{pedregosa2011scikit} or general frameworks like joblib\footnote{\url{https://github.com/joblib/joblib}}, may simply split the models into 4 subgroups by order and schedule the first 25 models (all $k$NNs) on worker 1, the next 25 models on worker 2, etc. This does not account for the fact that within a group of heterogeneous models, the computational cost varies. Scheduling the task with the equal number of models can result in highly imbalanced load. In the worst-case scenario, one worker may be assigned significant more load than the rest, resulting in halt to the entire process. In this example, the $k$NN subgroup will be the system curb due to high time complexity. 
% Obviously, this problem applies to both training and prediction stage.
One naive solution is to shuffle the base models randomly. However, there is no guarantee this heuristic could work, and it may be practically infeasible.

% If there are multiple workers available for distributed computing, BPS can assign tasks more evenly across all workers by forecasting model cost, for both training and prediction. 

\textbf{The proposed BPS focuses on delivering a more balanced task scheduling among workers via forecasting their cost in advance}. Ideally, all workers can finish the scheduled tasks within a similar duration and return the results. To achieve this goal, \system comes with a \textit{model cost predictor} $C_{\text{cost}}$ to forecast the model execution time (sum of 10 trials) given the meta-features (descriptive features) of a dataset \cite{zhao2020automating}.
% , including input data size, input data dimension, the algorithm embedding, etc.
$C_{\text{cost}}$ is trained on 11 algorithm family with 47 benchmark datasets by 10-fold cross validation, yielding an effective regressor (random forest is used in this study). $C_{\text{cost}}$'s outputs show high Spearman's Rank correlation \cite{spearman1904proof} ($r_s>0.9$)  to the true model cost rank with low p-value ($p<0.0001$), in all folds. Given a dataset and a model, $C_{\text{cost}}$ can predict the execution time of the model's execution time with high accuracy. It is noted that the model cost predictor is only trained for the major methods in Python Outlier Detection Toolbox (PyOD) \cite{zhao2019pyod}. For unseen models, they are classified as ``unknown" to be assigned with the max cost to prevent overoptimistic scheduling.

As a result, a scheduling strategy is proposed by enforcing a nearly equal rank sum by the forecasted execution time among all available workers. Fig. \ref{fig:balanceLearning} provides a simple example of scheduling $m$ models to 4 workers. More formally, before scheduling $m$ models for training (or prediction), cost predictor $C_{\text{cost}}$ is invoked to forecast the execution time of each model $M$ in $\mathcal{M}$ as $C_{\text{cost}}(M)$ and output the model cost rank of each model in $\{1,m\}$ (the higher the rank, the longer the forecasted execution time). If there are $t$ cores (workers), each worker will be assigned a group of models to achieve the objective of minimizing the taskload imbalance among workers (Eq. \ref{eq:balanced_learning_goal}). 

\begin{equation}
\label{eq:balanced_learning_goal}
\scriptsize
    \min_{\mathcal{W}} \sum_{i=1}^{t} \abs{\sum_{M_j \in \mathcal{W}_i}{C_{\text{cost}}(M_j)-\frac{m^2+m}{2t}}}
\end{equation}

Consequently, each worker is assigned with a group of models with the rank sum close to the average rank sum $\frac{(1+m)m}{2}/t=\frac{m^2+m}{2t}$. Indeed, the accurate running time prediction is less relevant as it depends on the hardware---the rank is more useful as a relevance measure with the transferability to other hardware. That is, the running time will vary on different machines, but the relative rank should preserve. One issue around the sum of ranks is the overestimation of high-rank models. For instance, rank $f$-th model will be counted $f$ times more heavily than rank $1$ model during the sum calculation, even their actual running time difference will not be as big as $f$ times. To fix this, we introduce a discounted rank by rescaling model rank $f$ to $1+\frac{\alpha f}{m}$, where $\alpha$ denotes the scaling strength (default to 1). A larger $\alpha$ therefore puts more emphasis on costly models.

\begin{table*}[!htb]
    \caption{Comparison of various data compression methods on different outlier detectors and datasets (see Appendix Table \ref{appendix:datasets}).
    Each column corresponds to an evaluation metric (execution time is measured in seconds); the best performing method is indicated in \textbf{bold}. JL projection methods, especially \textit{circulant} and \textit{toeplitz}, outperform regarding both time cost and prediction accuracy.}
    \footnotesize
    \label{table:projection_comparison}
   \begin{subtable}[t]{.3\textwidth}
        \centering
        \caption{ABOD on \textbf{Cardio}}\label{subtab:abod_cardio}
                \vspace{-0.05in}
            \begin{tabular}{@{\extracolsep{1pt}}l  r r r r }
                \toprule
                \textbf{Method}	&	\textbf{Time}	&	\textbf{ROC}		&   \textbf{P@N}     \\
                \midrule 
                original	&	0.98	&	0.59	&	0.25	\\
                PCA	&	\textbf{0.82}	&	0.59	&	0.26	\\
                RS &	0.92	&	\textbf{0.63}	&	\textbf{0.29}	\\
                \textit{basic}	&	0.83	&	0.62	&	0.28	\\
                \textit{discrete}	&	\textbf{0.82}	&	0.62	&	0.28	\\
                \textit{circulant}	&	0.83	&	0.62	&	0.27	\\
                \textit{toeplitz}	&	0.83	&	0.62	&	0.28	\\
                \bottomrule \\
            \end{tabular}
    \end{subtable}
    \hspace{\fill}
   \begin{subtable}[t]{.3\textwidth}
        \centering
        \caption{LOF on \textbf{Cardio}}\label{subtab:lof_cardio}
                \vspace{-0.05in}
            \begin{tabular}{@{\extracolsep{1pt}}l  r r r r }
                \toprule
                \textbf{Method}	&	\textbf{Time}	&	\textbf{ROC}		&   \textbf{P@N}     \\
                \midrule 
                original	&	0.08	&	0.55	&	0.17	\\
                PCA	&	\textbf{0.04}	&	0.56	&	0.19	\\
                RS &	\textbf{0.04}	&	0.57	&	0.15	\\
                \textit{basic}	&	\textbf{0.04}	&	\textbf{0.60}	&	0.20	\\
                \textit{discrete}	&	\textbf{0.04}	&	0.59	&	0.19	\\
                \textit{circulant}	&	\textbf{0.04}	&	0.59	&	0.20	\\
                \textit{toeplitz}	&	\textbf{0.04}	&	\textbf{0.60}	&	\textbf{0.21}	\\
                \bottomrule \\
            \end{tabular}
    \end{subtable}
   \hspace{\fill}
   \begin{subtable}[t]{.3\textwidth}
        \centering
        \caption{$k$NN on \textbf{Cardio}}\label{subtab:knn_cardio}
                \vspace{-0.05in}
            \begin{tabular}{@{\extracolsep{1pt}}l  r r r r }
                \toprule
                \textbf{Method}	&	\textbf{Time}	&	\textbf{ROC}		&   \textbf{P@N}     \\
                \midrule 
                original	&	0.09	&	0.71	&	0.34	\\
                PCA	&	\textbf{0.03}	&	0.73	&	0.34	\\
                RS &	\textbf{0.03}	&	0.69	&	\textbf{0.38}	\\
                \textit{basic}	&	\textbf{0.03}	&	\textbf{0.74}	&	0.35	\\
                \textit{discrete}	&	\textbf{0.03}	&	\textbf{0.74}	&	0.37	\\
                \textit{circulant}	&	\textbf{0.03}	&	\textbf{0.74}	&	0.34	\\
                \textit{toeplitz}	&	\textbf{0.03}	&	0.73	&	0.35	\\
                \bottomrule \\
            \end{tabular}
    \end{subtable}
    
    \begin{subtable}[t]{.3\textwidth}
        \centering
        \caption{ABOD on \textbf{MNIST}}\label{subtab:abod_mnist}
        \vspace{-0.05in}
            \begin{tabular}{@{\extracolsep{1pt}}l  r r r r }
                \toprule
                \textbf{Method}	&	\textbf{Time}	&	\textbf{ROC}		&   \textbf{P@N}     \\
                \midrule 
                original	        &	12.89	&	0.80	&	\textbf{0.39}	\\
                PCA	                &	8.93	&	\textbf{0.81}	&	0.37	\\
                RS	                &	\textbf{8.27}	&	0.74	&	0.32	\\
                \textit{basic}	    &	8.94	&	0.80	&	0.38	\\
                \textit{discrete}	&	8.86	&	0.80	&	\textbf{0.39}	\\
                \textit{circulant}	&	9.33	&	0.80	&	0.38	\\
                \textit{toeplitz}	&	8.96	&	0.80	&	0.38	\\
                \bottomrule \\
            \end{tabular}
    \end{subtable}%
    \hspace{\fill}
   \begin{subtable}[t]{.3\textwidth}
        \centering
        \caption{LOF on \textbf{MNIST}}\label{subtab:lof_mnist}
                \vspace{-0.05in}
            \begin{tabular}{@{\extracolsep{1pt}}l  r r r r }
                \toprule
                \textbf{Method}	&	\textbf{Time}	&	\textbf{ROC}		&   \textbf{P@N}     \\
                \midrule 
                original	&	7.64	&	0.68	&	0.29	\\
                PCA	&	4.92	&	0.67	&	0.27	\\
                RS &	\textbf{3.65}	&	0.63	&	0.23	\\
                \textit{basic}	&	4.87	&	0.70	&	0.31	\\
                \textit{discrete}	&	5.21	&	0.70	&	\textbf{0.32}	\\
                \textit{circulant}	&	5.06	&	0.69	&	0.31	\\
                \textit{toeplitz}	&	4.97	&	\textbf{0.71}	&	0.31	\\
                \bottomrule \\
            \end{tabular}
   \end{subtable}%
   \hspace{\fill}
   \begin{subtable}[t]{.3\textwidth}
        \centering
        \caption{$k$NN on \textbf{MNIST}}\label{subtab:knn_mnist}
                \vspace{-0.05in}
            \begin{tabular}{@{\extracolsep{1pt}}l  r r r r }
                \toprule
                \textbf{Method}	&	\textbf{Time}	&	\textbf{ROC}		&   \textbf{P@N}     \\
                \midrule 
                original	&	7.13	&	\textbf{0.84}	&	\textbf{0.42}	\\
                PCA	&	    3.92	&	\textbf{0.84}	&	0.40	\\
                RS &	\textbf{3.33}	&	0.77	&	0.34	\\
                \textit{basic}	&	4.17	&	\textbf{0.84}	&	\textbf{0.42}	\\
                \textit{discrete}	&	4.11	&	\textbf{0.84}	&	0.41	\\
                \textit{circulant}	&	4.13	&	\textbf{0.84}	&	0.41	\\
                \textit{toeplitz}	&	4.11	&	\textbf{0.84}	&	\textbf{0.42}	\\
                \bottomrule \\
            \end{tabular}
    \end{subtable}%
    
   \begin{subtable}[t]{.3\textwidth}
        \centering
        \caption{ABOD on \textbf{Satellite}}\label{subtab:abod_satellite}
                \vspace{-0.05in}
            \begin{tabular}{@{\extracolsep{1pt}}l  r r r r }
                \toprule
                \textbf{Method}	&	\textbf{Time}	&	\textbf{ROC}		&   \textbf{P@N}     \\
                \midrule 
                original	&	4.03	&	0.59	&	0.41	\\
                PCA	&	\textbf{3.01}	&	0.62	&	0.44	\\
                RS &	3.53	&	0.63	&	0.44	\\
                \textit{basic}	&	3.10	&	0.64	&	0.45	\\
                \textit{discrete}	&	3.12	&	0.65	&	0.46	\\
                \textit{circulant}	&	3.14	&	\textbf{0.66}	&	\textbf{0.48}	\\
                \textit{toeplitz}	&	3.14	&	\textbf{0.66}	&	0.47	\\
                \bottomrule \\
            \end{tabular}
    \end{subtable}
    \hspace{\fill}
   \begin{subtable}[t]{.3\textwidth}
        \centering
        \caption{LOF on \textbf{Satellite}}\label{subtab:lof_satellite}
                \vspace{-0.05in}
            \begin{tabular}{@{\extracolsep{1pt}}l  r r r r }
                \toprule
                \textbf{Method}	&	\textbf{Time}	&	\textbf{ROC}		&   \textbf{P@N}     \\
                \midrule 
                original	&	0.82	&	\textbf{0.55}	&	0.37	\\
                PCA	&	\textbf{0.23}	&	0.54	&	0.36	\\
                RS &	0.39	&	0.54	&	0.37	\\
                \textit{basic}	&	0.31	&	0.54	&	0.37	\\
                \textit{discrete}	&	0.32	&	0.54	&	0.37	\\
                \textit{circulant}	&	0.39	&	\textbf{0.55}	&	\textbf{0.38}	\\
                \textit{toeplitz}	&	0.37	&	0.54	&	0.37	\\
                \bottomrule \\
            \end{tabular}
    \end{subtable}
    \hspace{\fill}
   \begin{subtable}[t]{.3\textwidth}
        \centering
        \caption{$k$NN on \textbf{Satellite}}\label{subtab:knn_satellite}
                \vspace{-0.05in}
            \begin{tabular}{@{\extracolsep{1pt}}l  r r r r }
                \toprule
                \textbf{Method}	&	\textbf{Time}	&	\textbf{ROC}		&   \textbf{P@N}     \\
                \midrule 
                original	&	0.71	&	0.67	&	0.49	\\
                PCA	&	\textbf{0.18}	&	0.67	&	0.50	\\
                RS &	0.31	&	0.68	&	0.49	\\
                \textit{basic}	&	0.24	&	0.68	&	0.49	\\
                \textit{discrete}	&	0.25	&	0.69	&	0.50	\\
                \textit{circulant}	&	0.33	&	\textbf{0.70}	&	0.50	\\
                \textit{toeplitz}	&	0.30	&	\textbf{0.70}	&	\textbf{0.51}	\\
                \bottomrule \\
            \end{tabular}
    \end{subtable}
    
   \begin{subtable}[t]{.3\textwidth}
        \centering
        \caption{ABOD on \textbf{Satimage-2}}\label{subtab:abod_satimage}
                \vspace{-0.05in}
            \begin{tabular}{@{\extracolsep{1pt}}l  r r r r }
                \toprule
                \textbf{Method}	&	\textbf{Time}	&	\textbf{ROC}		&   \textbf{P@N}     \\
                \midrule 
                original	&	3.68	&	0.85	&	0.28	\\
                PCA	&	\textbf{2.70}	&	0.88	&	0.30	\\
                RS &	3.20	&	0.89	&	0.28	\\
                \textit{basic}	&	2.78	&	0.91	&	0.29	\\
                \textit{discrete}	&	2.79	&	0.91	&	\textbf{0.31}	\\
                \textit{circulant}	&	2.85	&	0.91	&	0.29	\\
                \textit{toeplitz}	&	2.83	&	\textbf{0.92}	&	0.30	\\
                \bottomrule \\
            \end{tabular}
    \end{subtable}
       \hspace{\fill}
   \begin{subtable}[t]{.3\textwidth}
        \centering
        \caption{LOF on \textbf{Satimage-2}}\label{subtab:lof_satimage}
                \vspace{-0.05in}
            \begin{tabular}{@{\extracolsep{1pt}}l  r r r r }
                \toprule
                \textbf{Method}	&	\textbf{Time}	&	\textbf{ROC}		&   \textbf{P@N}     \\
                \midrule 
                original	&	0.79	&	0.54	&	0.07	\\
                PCA	&	\textbf{0.20}	&	0.52	&	0.04	\\
                RS &	0.37	&	0.53	&	0.08	\\
                \textit{basic}	&	0.29	&	0.52	&	0.08	\\
                \textit{discrete}	&	0.30	&	0.53	&	0.07	\\
                \textit{circulant}	&	0.43	&	\textbf{0.59}	&	\textbf{0.11}	\\
                \textit{toeplitz}	&	0.32	&	0.54	&	0.09	\\
                \bottomrule \\
            \end{tabular}
    \end{subtable}
    \hspace{\fill}
   \begin{subtable}[t]{.3\textwidth}
        \centering
        \caption{$k$NN on \textbf{Satimage-2}}\label{subtab:knn_satimage}
                \vspace{-0.05in}
            \begin{tabular}{@{\extracolsep{1pt}}l  r r r r }
                \toprule
                \textbf{Method}	&	\textbf{Time}	&	\textbf{ROC}		&   \textbf{P@N}     \\
                \midrule 
                original	&	0.68	&	0.94	&	\textbf{0.39}	\\
                PCA	&	\textbf{0.15}	&	0.94	&	\textbf{0.39}	\\
                RS &	0.29	&	0.94	&	0.38	\\
                \textit{basic}	&	0.23	&	0.94	&	0.38	\\
                \textit{discrete}	&	0.20	&	0.95	&	0.37	\\
                \textit{circulant}	&	0.36	&	\textbf{0.96}	&	0.37	\\
                \textit{toeplitz}	&	0.25	&	\textbf{0.96}	&	\textbf{0.39}	\\
                \bottomrule \\
            \end{tabular}
    \end{subtable}
\end{table*}

\section{Experiments \& Discussion}
First, three experiments are conducted to understand the effectiveness of individual modules independently: \textbf{Q1}: how will different compression methods affect the performance of downstream OD accuracy (\S \ref{EXP:RP}); \textbf{Q2}: will use pseudo-supervised regressors lead to more efficient prediction in comparison to the original unsupervised models (\S \ref{EXP:PSA}) and \textbf{Q3}: how does the proposed balanced scheduling strategy perform under different settings (varying number of models $m$, number of workers $t$, etc.) (\S \ref{EXP:bps}). Then, the full \system with all three modules enabled is evaluated regarding time cost and prediction accuracy (on new samples) (\S \ref{EXP:full_system}). Finally, a real-world deployment case on fraudulent claim analysis at IQVIA (a leading healthcare organization), is described (\S \ref{EXP:deployment_Case}). 
The details of experiment setting, e.g., datasets, evaluation metrics, and the pool of heterogeneous OD models, can be found in the Appendix.

\subsection{Q1: Comparison of Model Compression Methods}
\label{EXP:RP}
In this section, \textbf{we demonstrate the effectiveness of RP module in high-dimensional OD tasks. }
To evaluate the effect of data projection, we choose three costly outlier detection algorithms, ABOD, LOF, and $k$NN to measure their execution time, and prediction accuracy (ROC and P@N), before and after projection. These methods directly or indirectly measure sample similarity in Euclidean space, e.g., pairwise distance, which is prone to the curse of dimensionality, where data compression can help. 

Table \ref{table:projection_comparison} shows the comparison results on four datasets (see Appendix Table \ref{appendix:datasets}); the reduced dimension is set as $k=\frac{2}{3}d$ ($33\%$ compression). We compare the proposed four JL projection methods (see \S \ref{sec:RP} for details of \textit{basic}, \textit{discrete}, \textit{circulant}, and \textit{toeplitz}) with \textbf{original} (no projection), \textbf{PCA}, and \textbf{RS} (randomly select $k$ features from the original $d$ features, used in Feature Bagging \cite{lazarevic2005feature} and LSCP \cite{zhao2019lscp}). First, all compression methods show superiority regarding time cost. Second, 
\textbf{original} (no compression) method
rarely outperforms, possibly due to the curse of dimensionality and lack of diversity \cite{zhao2019lscp}. 
Third, \textbf{PCA} is inferior to \textbf{original} regarding prediction accuracy (see LOF performance in Table \ref{subtab:lof_mnist}, \ref{subtab:lof_satellite}, and \ref{subtab:lof_satimage}). The observation supports our claim that PCA is not suited in this scenario (see \S \ref{subsec:scalability_issues}).
% Third, PCA is always slightly faster than JL projection methods, although the detector performance is inferior to \textbf{original} when PCA projection is used along with LOF (Table \ref{subtab:lof_mnist}, \ref{subtab:lof_satellite}, and \ref{subtab:lof_satimage}). The observation supports our claim that PCA is not suited in this scenario (see \S \ref{subsec:scalability_issues}).
Fourth, JL methods generally lead to equivalent or better prediction performance than  \textbf{original} regarding both time and prediction accuracy. Lastly, among all JL methods, \textit{circulant} and \textit{toeplitz} generally outperform others. For instance, \textit{toeplitz} brings more than 60\% time reduction for $k$NN, demonstrating the effectiveness of RP and chosen as the default choice.

\begin{figure*}[t]
\centering
    \includegraphics[width=0.95\linewidth]{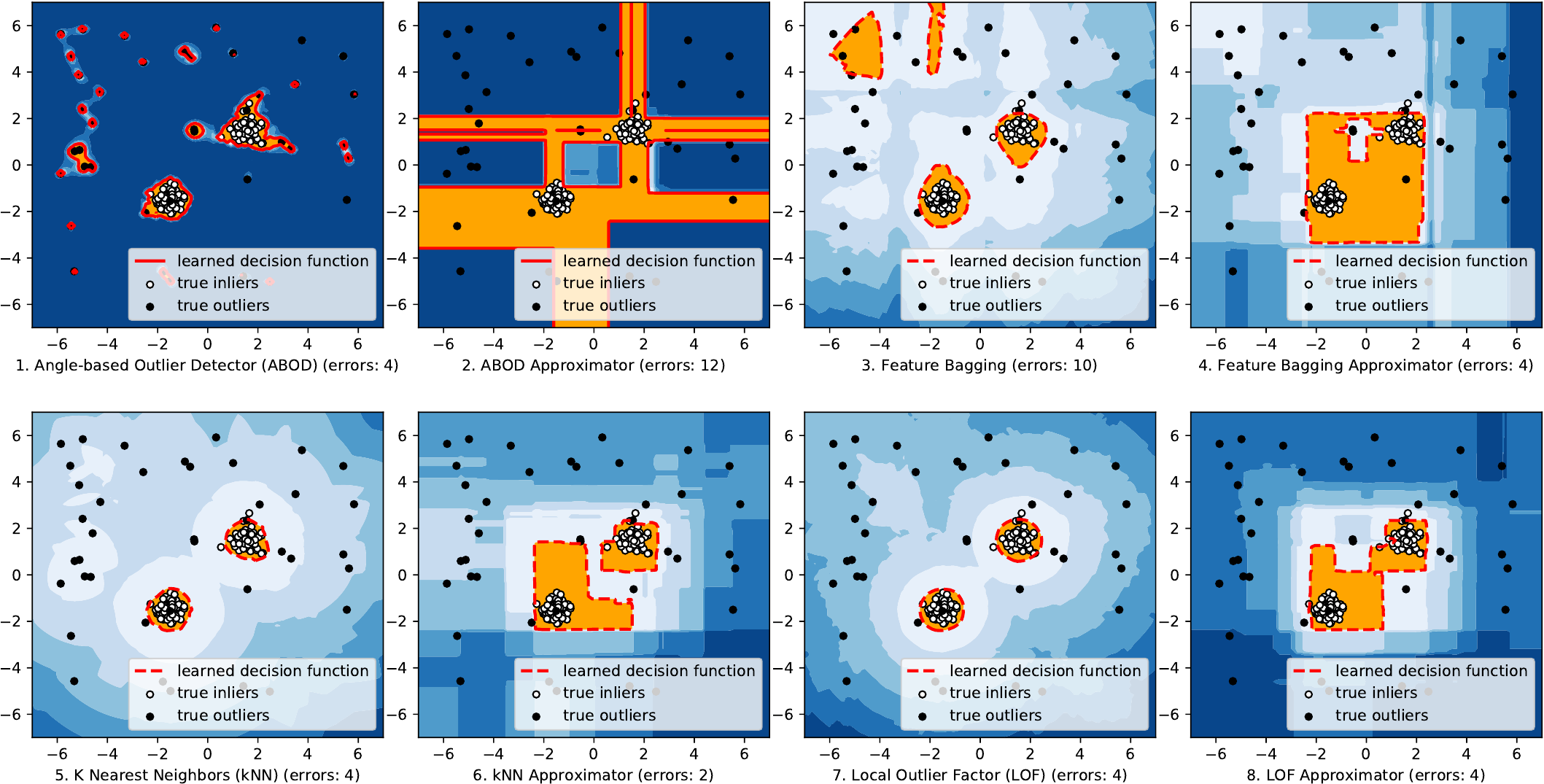}
\caption{Decision surface comparison among unsupervised models and their pseudo-supervised approximators (in pairs). The approximator's decision boundary shows a tentative regularization effect, leading to even fewer detection errors.} 
\label{fig:fig_all_comparision}
\end{figure*} 
\subsection{Q2: The Visual and Quantitative Analysis of PSA}
\label{EXP:PSA}
\textbf{Through  both visualization and quantitative analysis, we observe PSA is useful for accelerating prediction of proximity-based OD algorithms.}
To better understand the effect of pseudo-supervised approximation, we first generate a synthetic dataset with 200 two-dimensional samples, 
consisting of 40 outliers generated by Normal distribution and 160 normal samples generated from Uniform distribution. In Fig. \ref{fig:fig_all_comparision}, we plot the decision surfaces of four costly unsupervised models (ABOD, Feature Bagging, $k$NN, and LOF) and of their corresponding supervised approximators (random forest regressor), with accuracy errors reported. In general, the faster pseudo supervised approximators do not lead to more errors, justifying the effectiveness of approximation. Fig. \ref{fig:fig_all_comparision} subfigure 4 and 6 show that the approximators have even lower errors than the original (Feature Bagging and $k$NN). 
% It is noted that the decision surfaces of the approximators are different and some regularization effect appears. 
% We also notice that the decision surfaces are quite different from the original OD models and their approximators.
With a closer look at the decision surfaces, we assume that the approximation process improves the generalization ability of the model by ``ignoring" some  overfitted points. However, the approximation does not work with ABOD, possibly due to its extremely coarse decision surface (see Fig. \ref{fig:fig_all_comparision}, subfigure 1).

Table \ref{table:approx_roc} and \ref{table:approx_p@n} compare prediction performance (scoring on new-coming samples) between the original unsupervised models and pseudo-supervised approximators on 10 datasets with 6 costly algorithms, regarding ROC and P@N. Since these algorithms are more computationally expensive than random forest regressors for prediction (by time complexity analysis), we skip the prediction time comparison 
where the gain is clear. Consequently, the focus is whether the approximators could predict unseen samples as good as the original unsupervised models. 
% The acceptable threshold of performance degradation between an approximator and its original unsupervised models is set as $[0,0.01]$ and any negative difference larger than $0.01$ will be regarded as degradation. 
The tables reveal that not all the algorithms are suited for PSA, which is in line with the visual analysis. For instance, ABOD shows performance decrease on half of the datasets. Notably, ABOD looks for a low-dimensional subspace to embed the normal samples \cite{aggarwal2016outlier}, leading to a complex decision surface to approximate. In contrast, proximity-based models benefit from the approximation. Both tables show, $k$NN, LoF, and a$k$NN (average $k$NN) experience a performance gain. Specifically, all three algorithms yield around 100\% ROC increase on \textbf{HTTP}. Other algorithms, such as Feature Bagging and CBLOF, show a minor performance variation within the acceptable range. In other words, it is useful to perform PSA for these estimators as the time efficiency is greatly improved with little to no loss in prediction accuracy. 

\begin{table*}[ht]\centering
	\caption{Prediction ROC scores of unsupervised models (Orig) and their pseudo-supervised approximators (Appr) by the average of 10 independent trials. The better method within each pair is indicated in \textbf{bold}. The approximators (Appr) outperform in most cases.} % title of Table
	\label{table:approx_roc}
	\scriptsize
	\resizebox{\textwidth}{!}{% use resizebox with textwidth
	\begin{tabular}{l |r r| r r| r r| r r| r r| r r| r r| r r| r r| r r} % centered columns (12 columns)
		\toprule %inserts double horizontal lines
		\textbf{Dataset} & \multicolumn{2}{c|}{\textbf{Annthyroid}} & \multicolumn{2}{c|}{\textbf{Breastw}} & \multicolumn{2}{c|}{\textbf{Cardio}} & \multicolumn{2}{c|}{\textbf{HTTP}} &  \multicolumn{2}{c|}{\textbf{MNIST}} & \multicolumn{2}{c|}{\textbf{Pendigits}} & \multicolumn{2}{c|}{\textbf{Pima}} & \multicolumn{2}{c|}{\textbf{Satellite}} & \multicolumn{2}{c|}{\textbf{Satimage-2}} & \multicolumn{2}{c}{\textbf{Thyroid}}\\
		\midrule
		\textbf{Model} & Orig & Appr & Orig & Appr & Orig & Appr & Orig & Appr & Orig & Appr & Orig & Appr & Orig & Appr & Orig & Appr & Orig & Appr & Orig & Appr\\
		\midrule
        ABOD	&	\textbf{0.83}	&	0.71	&	0.92	&	\textbf{0.93}	&	\textbf{0.63}	&	0.53	&	\textbf{0.15}	&	0.13	&	\textbf{0.81}	&	0.79	&	0.67	&	\textbf{0.82}	&	0.66	&	\textbf{0.70}	&	0.59	&	\textbf{0.68} &	0.89	& \textbf{0.99} & \textbf{0.96} &	0.67\\
        CBLOF	&	0.67	&	\textbf{0.68}	&	0.96	&	\textbf{0.98}	&	0.73	&	\textbf{0.76}	&	\textbf{1.00}	&	\textbf{1.00}	&	0.85	&	\textbf{0.89}	&	\textbf{0.93}	&	\textbf{0.93}	&	0.63	&	\textbf{0.68}	&	0.72	&	\textbf{0.77}    &	\textbf{1.00}	&   \textbf{1.00} & 0.92	& \textbf{0.97}\\
        FB	&	\textbf{0.81}	&	0.45	&	\textbf{0.34}	&	0.10	&	0.61	&	\textbf{0.70}	&	0.34	&	\textbf{0.97}	&	0.72	&	\textbf{0.83}	&	0.39	&	\textbf{0.51}	&	0.59	&	\textbf{0.63}	&	0.53	&	\textbf{0.64}  & 0.36 & \textbf{0.40}	& \textbf{0.83}	& 0.46\\
        % HBOS	&	0.62	&	0.62	&	0.98	&	0.98	&	0.83	&	0.83	&	0.99	&	0.99	&	0.60	&	0.65	&	0.93	&	0.95	&	0.71	&	0.73	&	0.73	&	0.74	& 0.97	& 0.98  & 0.95	& 0.95\\
        % IF	&	0.82	&	0.82	&	0.98	&	0.98	&	0.93	&	0.93	&	1.00	&	1.00	&	0.80	&	0.82	&	0.94	&	0.94	&	0.63	&	0.67	&	0.69	&	0.70	&  0.99	    &   0.99 & 0.98	& 0.99\\
        $k$NN	&	\textbf{0.80}	&	0.79	&	\textbf{0.97}	&	\textbf{0.97}	&	0.73	&	\textbf{0.75}	&	0.19	&	\textbf{0.85}	&	0.85	&	\textbf{0.86}	&	0.74	&	\textbf{0.87}	&	0.69	&	\textbf{0.71}	&	0.68	&	\textbf{0.75}	& 0.96	&   \textbf{0.99} & 0.97	& \textbf{0.98}\\
        a$k$NN	&	0.81	&	\textbf{0.82}	&	\textbf{0.97}	&	\textbf{0.97}	&	0.67	&	\textbf{0.72}	&	0.19	&	\textbf{0.88}	&	0.84	&	\textbf{0.85}	&	0.72	&	\textbf{0.87}	&	0.69	&	\textbf{0.71}	&	0.66	&	\textbf{0.74}	& 0.95	& \textbf{0.99} & 0.97	& \textbf{0.98}\\
        LOF	&	0.74	&	\textbf{0.85}	&	0.44	&	\textbf{0.45}	&	0.60	&	\textbf{0.68}	&	0.35	&	\textbf{0.75}	&	0.72	&	\textbf{0.76}	&	0.38	&	\textbf{0.47}	&	0.59	&	\textbf{0.65}	&	0.53	&	\textbf{0.66}	& 0.36	&\textbf{0.38} & 0.80 & \textbf{0.95}\\
        % MCD}}	&	0.92	&	0.85}}	&	0.98	&	0.98	&	0.84	&	0.79}}	&	1.00	&	1.00	&	0.88	&	0.81}}	&	0.82	&	0.85	&	0.65	&	0.68	&	0.81	&	0.85	& 0.99	& 1.00 & 0.98 & 0.99\\
        % OCSVM	&	0.68	&	0.67	&	0.96	&	0.97	&	0.95	&	0.95	&	0.99	&	1.00	&	0.86	&	0.89	&	0.93	&	0.93	&	0.58	&	0.61	&	0.66	&	0.67  & 0.99 &	0.99 & 0.97 & 0.97\\
        % PCA}}	&	0.68	&	0.67	&	0.96	&	0.96	&	0.96	&	0.50}}	&	1.00	&	1.00	&	0.87	&	0.50}}	&	0.93	&	0.93	&	0.59	&	0.61	&	0.60	&	0.60  & 0.97 &	0.97 & 0.96 & 0.97\\
    \midrule %inserts single line
	\end{tabular}}
\end{table*}

\begin{table*}[ht]\centering
	\caption{Prediction P@N scores of unsupervised models (Orig) and their pseudo-supervised approximators (Appr) by the average of 10 independent trials. The better method within each pair is indicated in \textbf{bold}. The approximators (Appr) outperform in most cases.} % title of Table
	\label{table:approx_p@n}
	\scriptsize
	\resizebox{\textwidth}{!}{% use resizebox with textwidth
	\begin{tabular}{l |r r| r r| r r| r r| r r| r r| r r| r r| r r| r r} % centered columns (12 columns)
		\toprule %inserts double horizontal lines
		\textbf{Dataset} & \multicolumn{2}{c|}{\textbf{Annthyroid}} & \multicolumn{2}{c|}{\textbf{Breastw}} & \multicolumn{2}{c|}{\textbf{Cardio}} & \multicolumn{2}{c|}{\textbf{HTTP}} &  \multicolumn{2}{c|}{\textbf{MNIST}} & \multicolumn{2}{c|}{\textbf{Pendigits}} & \multicolumn{2}{c|}{\textbf{Pima}} & \multicolumn{2}{c|}{\textbf{Satellite}} & \multicolumn{2}{c|}{\textbf{Satimage-2}} & \multicolumn{2}{c}{\textbf{Thyroid}}\\
		\midrule
		\textbf{Model} & Orig & Appr & Orig & Appr & Orig & Appr & Orig & Appr & Orig & Appr & Orig & Appr & Orig & Appr & Orig & Appr & Orig & Appr & Orig & Appr\\
        ABOD	&	\textbf{0.31}	&	0.08	&	0.80	&	\textbf{0.83}	&	\textbf{0.27}	&	0.20	&	0.00	&	0.00	&	\textbf{0.40}	&	0.27	&	\textbf{0.05}	&	\textbf{0.05}	&	0.48	&	\textbf{0.52}	&	0.41	&	\textbf{0.46}	&	0.21	&	\textbf{0.64}	&	\textbf{0.36}	&	0.00\\
        CBLOF	&	\textbf{0.25}	&	0.24	&	0.86	&	\textbf{0.90}	&	0.31	&	\textbf{0.34}	&	\textbf{0.02}	&	0.01	&	0.42	&	\textbf{0.48}	&	0.35	&	\textbf{0.36}	&	0.43	&	\textbf{0.48}	&	0.54	&	\textbf{0.57} &	\textbf{0.96}	&	\textbf{0.96}	&	0.26	&	\textbf{0.38}	\\
        FB	&	\textbf{0.24}	&	0.02	&	0.03	&	\textbf{0.07}	&	0.23	&	\textbf{0.26}	&	0.02	&	\textbf{0.04}	&	0.34	&	\textbf{0.36}	&	0.03	&	\textbf{0.07}	&	0.37	&	\textbf{0.44}	&	0.37	&	\textbf{0.42} &	0.03	&	\textbf{0.04}	&	\textbf{0.05}	&	0.02\\
        % HBOS	&	0.28	&	0.27	&	0.92	&	0.92	&	0.50	&	0.50	&	0.02	&	0.02	&	0.13	&	0.14	&	0.32	&	0.35	&	0.52	&	0.51	&	0.56	&	0.55 &	0.7857	&	0.8214	&	0.4722	&	0.4722	\\
        % IF	&	0.31	&	0.30	&	0.90	&	0.90	&	0.49	&	0.55	&	0.85	&	0.88	&	0.29	&	0.31	&	0.40	&	0.40	&	0.44	&	0.47	&	0.55	&	0.58 &	0.95002	&	0.92145	&	0.58332	&	0.58609	\\
        $k$NN	&	0.30	&	\textbf{0.32}	&	\textbf{0.89}	&	\textbf{0.89}	&	0.37	&	\textbf{0.46}	&	\textbf{0.03}	&	\textbf{0.03}	&	0.42	&	\textbf{0.45}	&	\textbf{0.08}	&	0.06	&	\textbf{0.47}	&	\textbf{0.47}	&	0.49	&	\textbf{0.53}	&	0.32	&	\textbf{0.43}	&	0.33	&	\textbf{0.42}\\
        A$k$NN	&	0.30	&	\textbf{0.33}	&	0.88	&	\textbf{0.89}	&	0.34	&	\textbf{0.40}	&	\textbf{0.03}	&	\textbf{0.03}	&	0.41	&	\textbf{0.45}	&	0.05	&	\textbf{0.13}	&	0.48	&	\textbf{0.49}	&	0.47	&	\textbf{0.52}  &	0.25	&	\textbf{0.43}	&	0.31	&	\textbf{0.44}	\\
        LOF	&	0.27	&	\textbf{0.36}	&	0.19	&	\textbf{0.35}	&	\textbf{0.23}	&	\textbf{0.23}	&	0.01	&	\textbf{0.03}	&	\textbf{0.33}	&	0.32	&	0.03	&	\textbf{0.08}	&	0.40	&	\textbf{0.44}	&	0.37	&	\textbf{0.42}	&	0.04	&	\textbf{0.07}	&	0.19	&	\textbf{0.25}\\
        % MCD}}	&	0.45	&	0.46	&	0.91	&	0.91	&	0.42	&	0.46	&	0.11	&	0.08}}	&	0.34	&	0.35	&	0.08	&	0.08	&	0.45	&	0.47	&	0.69	&	0.72	&	0.5	&	0.64287	&	0.6389	&	0.69067\\
        % OCSVM	&	0.25	&	0.24	&	0.89	&	0.92	&	0.51	&	0.57	&	0.02	&	0.02	&	0.41	&	0.42	&	0.37	&	0.41	&	0.40	&	0.43	&	0.53	&	0.56	&	0.9643	&	0.9643	&	0.4167	&	0.4444\\
        % PCA}}	&	0.23	&	0.23	&	0.91	&	0.93	&	0.59	&	0.00}}	&	0.02	&	0.02	&	0.40	&	0.00}}	&	0.32	&	0.33	&	0.40	&	0.43	&	0.48	&	0.47	&	0.8929	&	0.8929	&	0.4167	&	0.4167\\
    \bottomrule %inserts single line
	\end{tabular}}
\end{table*}

\begin{table}[!htb]
\centering
	\caption{Training time comparison (in seconds) between Simple scheduling and BPS against various number of OD models and workers.
	Percent of time reduction, Redu (\%), is indicated in \textbf{bold}. BPS consistently outperform to Generic scheduling}. % title of Table
	\scriptsize
	\begin{tabular}{l l  l | l  l |  r  r | r } % centered columns (4 columns)
		\toprule %inserts double horizontal lines
		\textbf{Dataset} & \textbf{\textit{n}} & \textbf{\textit{d}} & \textbf{\textit{m}} & \textbf{\textit{t}} & \textbf{Generic} & \textbf{BPS} & \textbf{Redu (\%)}\\
		\midrule
        Cardio & 1831	&	21	&	500	    &	2	&	240.12	&	221.34	&	\textbf{7.82} \\
        Cardio & 1831	&	21	&	500	    &	4	&	185.44	&	154.43	&	\textbf{16.72} \\
        Cardio & 1831	&	21	&	500	    &	8	&	140.63	&	120.02	&	\textbf{14.65} \\
        Cardio & 1831	&	21	&	1000	&	2	&	199.77	&	185.63	&	\textbf{7.08} \\
        Cardio & 1831	&	21	&	1000	&	4	&	130.82	&	110.60	&	\textbf{15.45} \\
        Cardio & 1831	&	21	&	1000	&	8	&	97.75	&	73.43	&	\textbf{24.88} \\
        \midrule
        Letter & 1600    &   32  &   500     &   2   &   111.95  &   109.52  &   \textbf{2.17}  \\
        Letter & 1600    &   32  &   500     &   4   &   92.69   &   86.24   &   \textbf{6.94}  \\
        Letter & 1600    &   32  &   500     &   8   &   57.21   &   48.72   &   \textbf{14.84} \\
        Letter & 1600    &   32  &   1000    &   2   &   224.61  &   222.59  &   \textbf{0.90}  \\
        Letter & 1600    &   32  &   1000    &   4   &   228.08  &   172.07  &   \textbf{24.56} \\
        Letter & 1600    &   32  &   1000    &   8   &   109.50  &   89.51   &   \textbf{17.80} \\
		\midrule
        PageBlock & 5393	&	10	&	100	&	2	&	51.11	&	35.17	&	\textbf{31.19} \\
        PageBlock & 5393	&	10	&	100	&	4	&	42.49	&	16.23	&	\textbf{61.80} \\
        PageBlock & 5393	&	10	&	100	&	8	&	38.45	&	16.97	&	\textbf{55.86} \\
        PageBlock & 5393	&	10	&	500	&	2	&	197.84	&	137.46	&	\textbf{30.52} \\
        PageBlock & 5393	&	10	&	500	&	4	&	167.36	&	76.14	&	\textbf{54.51} \\
        PageBlock & 5393	&	10	&	500	&	8	&	127.08	&	66.29	&	\textbf{47.84} \\
        \midrule
        Pendigits & 6870	&	16	&	500	    &	2	&	351.97	&	287.14	&	\textbf{18.42} \\
        Pendigits & 6870	&	16	&	500	    &	4	&	288.51	&	146.50	&	\textbf{49.22} \\
        Pendigits & 6870	&	16	&	500	    &	8	&	180.86	&	102.11	&	\textbf{43.33} \\
        Pendigits & 6870	&	16	&	1000	&	2	&	697.20	&	561.15	&	\textbf{19.51} \\
        Pendigits & 6870	&	16	&	1000	&	4	&	579.70	&	288.11	&	\textbf{50.33} \\
        Pendigits & 6870	&	16	&	1000	&	8	&	365.20	&	182.32	&	\textbf{50.08} \\
		\bottomrule
	\end{tabular}
	\label{table:BPS} % is used to refer this table in the text
\end{table}

\subsection{Q3: Time Reduction of Balanced Scheduling}
\label{EXP:bps}
To evaluate the effectiveness of the proposed BPS algorithm, we run the following experiments by varying: (i) the size ($n$) and the dimension ($d$) of the datasets, (ii) the number of estimators ($m$) and (iii) the number of CPU cores ($t$). Due to the space limit, we only show the training time comparison between the generic scheduling and BPS on \textbf{Cardio}, \textbf{Letter}, \textbf{PageBlock}, and \textbf{Pendigits}, by setting $m \in \{100,500\}$ and $t \in \{2,4,8\}$, consistent with the single machine setting in real-world applications.

% More results can be found on the online supplementary, and the conclusion holds for all tested datasets. 
Table \ref{table:BPS} shows that the proposed BPS has a clear edge over the generic scheduling mechanism (denoted as \textbf{Generic} in the tables) that equally splits the tasks by order. It yields a significant time reduction (denoted as \textbf{\% Redu} in the table), 
which gets more remarkable if more cores are used along with large datasets. For instance, the time reduction is more than 40\% on \textbf{PageBlock} and \textbf{Pendigits} when 8 cores are used. This agrees with our assumption that model cost vary more drastically on large datasets given the time complexity increase non-linearly to the size---the proposed BPS method is particularly helpful.

\begin{table*}[!htp]
\centering
	\caption{Comparison between the baseline (denoted as \_B) and \system (denoted as \_S) regarding time cost, and prediction accuracy (ROC and P@N). The better method within each pair is indicated in \textbf{bold} (Optdigits fail to yield meaningful P@N). \system generally brings time reduction with no loss in prediction accuracy on majority of datasets.} % title of Table
	\scriptsize
	\resizebox{0.95\textwidth}{!}{% use resizebox with textwidth
	\begin{tabular}{l l l l | r r | r r |  r r |  r r |  r r |  r r } % centered columns (4 columns)
		\toprule %inserts double horizontal lines
		\multicolumn{4}{c|}{\textbf{Data Information}} & \multicolumn{4}{c|}{\textbf{Time Cost (in seconds)}} & \multicolumn{4}{c|}{\textbf{Ensemble Model Performance (ROC)}} & \multicolumn{4}{c}{\textbf{Ensemble Model Performance (P@N)}}\\
		\midrule
		\textbf{Dataset} & \textbf{\textit{n}} & \textbf{\textit{d}} & \textbf{\textit{t}} & \textbf{Fit\_B} & \textbf{Fit\_S} & \textbf{Pred\_B} & \textbf{Pred\_S} & \textbf{Avg\_B} & \textbf{Avg\_S} & \textbf{MOA\_B} & \textbf{MOA\_S} & \textbf{Avg\_B} & \textbf{Avg\_S} & \textbf{MOA\_B} & \textbf{MOA\_S}\\
        \midrule
        Annthyroid  & 7200	&	6	&	5	&	73.91	&	\textbf{65.23}  &  47.48	&	\textbf{44.26} &  0.91	&	\textbf{0.93} &  0.91 &	\textbf{0.93}  & 0.46& \textbf{0.54}& 0.46& \textbf{0.55}
\\
        Annthyroid  & 7200	&	6	&	10	&	71.00	&	\textbf{42.94}  &  44.68	&	\textbf{38.66} &  0.91 &	\textbf{0.93} &  0.92 &	\textbf{0.93}  & 0.46& \textbf{0.54}& 0.46& \textbf{0.54}\\
        Annthyroid  & 7200	&	6	&	30	&	42.80	&	\textbf{33.98}  &  30.92	&	\textbf{25.67} &  0.91	&	\textbf{0.93} &  0.92 &	\textbf{0.93}  & 0.46& \textbf{0.54}& 0.46& \textbf{0.54}\\
        \midrule
        Cardio  & 1831	&	21	&	5	&	\textbf{78.84}	&	79.70  &  \textbf{46.09}	&	46.68 &  0.91	&	\textbf{0.93} &  0.91 &	\textbf{0.93}  & 0.46 & \textbf{0.54}& 0.45& \textbf{0.55}\\
		Cardio  & 1831	&	21	&	10	&	72.04	&	\textbf{53.43}  &  44.57	&	\textbf{38.31} &  0.91	&	\textbf{0.93} &  0.91 &	\textbf{0.93}  & 0.46 & \textbf{0.54}& 0.46& \textbf{0.54}\\
		Cardio  & 1831	&	21	&	30	&	47.53	&	\textbf{44.57}  &  \textbf{31.31} &	31.43 &  0.91	&	\textbf{0.93} &  0.92 &	\textbf{0.93}  & 0.46 & \textbf{0.54} & 0.46& \textbf{0.55}\\
		\midrule
		MNIST  & 7603	&	100	&	5	&	856.53	&	\textbf{748.40}  &  453.39	&	\textbf{324.76} &  0.77	&	\textbf{0.81} &  0.77	&	\textbf{0.81}  & 0.29& \textbf{0.35}& 0.28& \textbf{0.34}\\
		MNIST  & 7603	&	100	&	10	&	726.76	&	\textbf{573.66}  &  367.85	&	\textbf{328.95} &  0.78	    &	\textbf{0.81} &  0.78	&	\textbf{0.81}  & 0.29& \textbf{0.35}& 0.30& \textbf{0.34}\\
		MNIST  & 7603	&	100	&	30	&	357.40	&	\textbf{329.71}  &  260.80	&	\textbf{134.08} &  0.78	&	\textbf{0.81} &  0.78	&	\textbf{0.81}  & 0.29& \textbf{0.35}& 0.29& \textbf{0.34}\\
        \midrule
		Optdigits  & 5216	&	64	&	5	&	295.38	&	\textbf{267.71} &  162.28	&	\textbf{149.19} &  0.73	&	\textbf{0.75} &  0.75 &	\textbf{0.77}  & 0.00& 0.00& 0.00& 0.00\\
		Optdigits  & 5216	&	64	&	10	&	247.24	&	\textbf{224.82} &  136.12	&	\textbf{125.54} &  0.73	&	\textbf{0.75} &  0.74 &	\textbf{0.75}  & 0.00& 0.00& 0.00& 0.00\\
		Optdigits  & 5216	&	64	&	30	&	825.23	&	\textbf{791.95} &  110.06	&	\textbf{62.63}  &  0.73    &   \textbf{0.75} &  0.73 &  \textbf{0.76}  & 0.00& 0.00& 0.00& 0.00\\
		\midrule
		Pendigits  & 6870  & 16	&	5	&	287.75	&	\textbf{282.25}  &  184.20	&	\textbf{158.26} &  0.92	&	\textbf{0.95} &  0.92	&	\textbf{0.94}  & 0.19& \textbf{0.23}& 0.19& \textbf{0.20}\\
		Pendigits  & 6870  & 16	&	10	&	281.49	&	\textbf{155.06}  &  179.83	&	\textbf{160.94} &  0.92	&	\textbf{0.95} &  0.92	&	\textbf{0.94}  & 0.19& \textbf{0.25}& 0.19& \textbf{0.23}\\
		Pendigits  & 6870  & 16	&	30	&	149.93	&	\textbf{145.59}  &  104.25	&	\textbf{89.85}  &  0.92	&	\textbf{0.94} &  0.93	    &   \textbf{0.94}  & 0.19& \textbf{0.25}& 0.19& \textbf{0.22}\\
		\midrule
		Pima  & 768	&	8	&	5	&	\textbf{28.72}	&	31.94  &  \textbf{21.16}	&	23.79 &  \textbf{0.71}	&	\textbf{0.71} &  \textbf{0.71}	&	0.70  & \textbf{0.51}& \textbf{0.51}& \textbf{0.53}& 0.51\\
		Pima  & 768	&	8	&	10	&	27.38	&	\textbf{20.15}  &  \textbf{20.81}	&	25.03 &  \textbf{0.71}	&	0.70 &  \textbf{0.71}	&	0.70  & \textbf{0.51}& \textbf{0.51}& \textbf{0.51}& \textbf{0.51}\\
		Pima  & 768	&	8	&	30	&	19.36	&	\textbf{17.89}  &  \textbf{13.83}	&	17.43 &  \textbf{0.71}	&	0.70 &  \textbf{0.71}	&	0.70  & \textbf{0.51}& 0.50& \textbf{0.52}& 0.50\\
		\midrule
		Shuttle & 49097  & 9	&	5	&	3326.54	&	\textbf{1453.93}  &  2257.50	&	\textbf{1956.12} &  \textbf{0.99}	&	\textbf{0.99} &  \textbf{0.99}	&	\textbf{0.99}  & \textbf{0.95} & \textbf{0.95}& \textbf{0.95}& \textbf{0.95}\\
		Shuttle & 49097  & 9    &	10	&	2437.10	&	\textbf{1396.21}  &  1549.97	&	\textbf{1321.16} & \textbf{0.99}	&	\textbf{0.99} &  \textbf{0.99}	&	\textbf{0.99}  & \textbf{0.95} & \textbf{0.95}& \textbf{0.95}& \textbf{0.95}\\
		Shuttle & 49097  & 9	&	30	&	1378.29	&	\textbf{1258.69}  &  837.41	&	\textbf{651.00}  &  \textbf{0.99}	&	\textbf{0.99} &  \textbf{0.99}	&	\textbf{0.99}  & \textbf{0.95} & \textbf{0.95}& \textbf{0.95}& \textbf{0.95}\\
		\midrule
		SpamSpace  & 4207	&	57	&	5	&	247.98	&	\textbf{244.39}  &  130.95	&	\textbf{110.08} &  \textbf{0.57}	&	0.56 &  \textbf{0.56}	&	\textbf{0.56}  & \textbf{0.45}& \textbf{0.45}& \textbf{0.46}& 0.45\\
		SpamSpace  & 4207	&	57	&	10	&	233.39	&	\textbf{186.91}  &  128.24	&	\textbf{115.83} &  \textbf{0.57}	&	0.56 &  \textbf{0.56}	&	\textbf{0.56}  & \textbf{0.46}& 0.45& \textbf{0.46}& \textbf{0.46}\\
		SpamSpace  & 4207	&	57	&	30	&	604.00	&	\textbf{538.91}  &  70.19	&	\textbf{61.38}  &  \textbf{0.57}	&	0.56 &  \textbf{0.57}	&	0.56  & \textbf{0.46}& \textbf{0.46}& \textbf{0.46}& 0.45\\
		\midrule
		Thyroid  & 3772	&	6	&	5	&	87.90	&	\textbf{71.34}  &  49.51	&	\textbf{48.20} &  0.91	&	\textbf{0.93} &  0.91	&	\textbf{0.93}  & 0.46 & \textbf{0.54}& 0.46& \textbf{0.55}\\
		Thyroid  & 3772	&	6	&	10	&	74.76	&	\textbf{46.91}  &  44.81	&	\textbf{38.60} &  0.91	&	\textbf{0.93} &  0.91	&	\textbf{0.93}  & 0.46& \textbf{0.54}& 0.46& \textbf{0.54}\\
		Thyroid  & 3772	&	6	&	30	&	45.84	&	\textbf{43.86}  &  28.90	&	\textbf{26.75} &  0.91	&	\textbf{0.93} &  0.92	&	\textbf{0.93}  & 0.46& \textbf{0.54}& 0.46& \textbf{0.54}\\
		\midrule
		Waveform  & 3443	&	21	&	5	&	167.98	& \textbf{147.00}	& 109.94  &  \textbf{94.46}	 &	\textbf{0.78} &  0.76	& \textbf{0.78} &  0.76  & 0.11& \textbf{0.13}& 0.11& \textbf{0.13}\\
		Waveform  & 3443	&	21	&	10	&	154.72	& \textbf{94.36}     & 91.69   &	 \textbf{55.17}   &   \textbf{0.78} &  0.76    & \textbf{0.78} &	0.77  & \textbf{0.11}& \textbf{0.11}& \textbf{0.11}& \textbf{0.11}\\
		Waveform  & 3443	&	21	&	30	&	97.11	& \textbf{95.77}     &  53.47  &	\textbf{48.04}    &  \textbf{0.78}	&	0.76 &  \textbf{0.78} &	0.76  & 0.11& \textbf{0.13}& 0.11& \textbf{0.13}\\
        \bottomrule
	\end{tabular}}
	\label{table:system} % is used to refer this table in the text
\end{table*}
\subsection{Full System Evaluation with All Modules}
\label{EXP:full_system}
Table \ref{table:system} shows the performance of \system with all three modules enabled, even not all of them are always needed in practice. In total, 600 hundred randomly selected OD models from \texttt{PyOD} are trained and tested on 10 datasets. To simulate the ``worst-case scenario", we build the model pool $\mathcal{M}$ by randomly select OD models, which minimizes the intrinsic task load imbalance. In real-world applications, this order randomization may not be possible as discussed in \S \ref{sec:bps}---two adjacent models are often from the same algorithm family and more prone to scheduling imbalance. \textbf{Although this setting will make the effectiveness of BPS module less impressive, we choose it to provide an empirical worst-case performance guarantee---the framework should generally perform better in practice.} 
% The elapsed time for fit and prediction (scoring on new samples), test ROC and P@N are reported. 

\system consistently yields promising results even we deliberately choose the unfavored setting. \textbf{Fit\_B} and \textbf{Pred\_B} denote the fit and prediction time of the baseline setting (no compression, no approximation, generic parallel task scheduling; see \S \ref{subsec:scalability_issues}). In comparison, \system (denoted as \textbf{Fit\_S} and \textbf{Pred\_S}) brings time reduction on majority of the datasets with minor to no performance degradation. To measure the prediction performance, we measure the ROC and P@N by averaging the base model results (denoted as \textbf{Avg\_}) and the maximum of average of the base models (denoted as \textbf{MOA\_}), a widely used two-phase outlier score combination framework \cite{Aggarwal2017}. Surprisingly, \system even leads to small performance boost in scoring new samples on most of the datasets (\textbf{Annthyroid}, \textbf{Cardio}, \textbf{MNIST}, \textbf{Optdigits}, \textbf{Pendigits}, and \textbf{Thyroid}). This performance gain may be jointly credited to the regularization effect by the randomness injected in JL projection (\S \ref{sec:RP}) and the pseudo-approximation (\S \ref{sec:psa})---the baseline setting may be overfitted on certain datasets. It is noted that \system leads to more improvement on high-dimensional, large datasets. For instance, the fit time is significantly reduced on \textbf{Shuttle} (more than $50\%$). On the contrary, \system is less meaningful for small datasets like \textbf{Pima} and \textbf{Cardio}, although they may also yield performance improvement. Again, our settings mimics the worst case scenario for \system (the model order is already randomly shuffled) but still observe a great performance improvement; real-world applications should generally expect more significant results.  

% It is also interesting to notice for majority of the datatsets, the most significant time reduction happens with the number of cores $t=10$. Indeed, the absolute value of $t$ is less meaningful, as the effect also depends on the number of base models. In our experiment setting, $t=5$ discounts the effect of BPS because the model order is already randomized, while $t=30$ will not benefit too much from BPS as each worker only needs to handle 20 detectors. This suggests larger performance gain may be achieved by selecting an appropriate number of workers for the base models.

% we may assume a large number of distributed workers reduce the time cost of each worker and the cost of the model cost forecast $C_{\text{cost}}$ 

\subsection{Real-World Deployment: Fraudulent Medical Claim Analysis at IQVIA}
\label{EXP:deployment_Case}

Estimated by the United States Government Accountability Office and Federal Bureau of Investigation, healthcare frauds cost American taxpayers tens of billions dollars a year \cite{aldrich2014much,Bagdoyan2018much}. Detecting fraudulent medical claims is crucial for taxpayers, pharmaceutical companies and insurance companies. To further demonstrate \system's performance on industry data, we deploy it on a proprietary pharmacy claim dataset owned by IQVIA (a leading healthcare firm) consisting of 123,720 medical claims among which 19,033 (15.38\%) are labeled as fraudulent. In each of the claim, there are 35 features including information such as drug brand, copay amount, insurance details, location and pharmacy/patient demographics. The current system in use is based on a group of selected detection models in \texttt{PyOD}, and an averaging method is applied on top of the base model results as the initial result. The cases marked as high risk are then transferred to human investigators in special investigation unit (SIU) for verification. It is important to provide prompt and accurate first-round screening for SIU, which leads to huge expense save.

\system is applied on top of the aforementioned dataset (74,220 records are used for training and 49,500 records are set aside for validation). Similarly to the full framework evaluation in \S \ref{EXP:full_system}, the new system with \system (all three modules enabled) is compared with the current distributed system on 10 cores. The fit time is reduced from 6232.54 seconds to 4202.30 seconds (32.57\% reduction), and the prediction time is reduced from 3723.45 seconds reduced to 2814.92 seconds (24.40\%). In addition to the time reduction, ROC and P@N also show improvements at 3.59\% and 7.46\%, respectively. Through this case, we are confident the proposed framework can be useful for many real-world applications for scalable outlier detection.

\section{Conclusion \& Future Directions}

In this work, we propose \system to expedite the training and prediction of a large number of unsupervised heterogeneous outlier detection models. It consists of three modules with focus on different levels (data, model, execution): (i) Random Projection module compresses data into low-dimensional subspaces to alleviate the curse of dimensionality; 
% using Johnson-Lindenstrauss projection
(ii) Pseudo-supervised Approximation module could accelerate costly unsupervised models' prediction by replacing them by faster supervised regressors, which also brings the extra benefit, e.g., interpretability and (iii) Balanced Parallel Scheduling module ensures that nearly equal amount of workload is assigned to available workers in distributed computing. The extensive experiments on more than 20 benchmark datasets and a real-world claim fraud analysis case show the great potential of \system, and many intriguing results are observed. 
For reproducibility and accessibility, all code, figures, and datasets are openly shared\footnote{\url{https://github.com/yzhao062/SUOD}}. 
% By the submission time, the \system has been widely used by practitioners with more than 700,000 downloads.
% All code, figures, implementation for demo and production, is released\footnote{\url{https://github.com/yzhao062/SUOD}} for model reproducibility and accessibility.

% cc: Many investigations are underway
More investigations are currently underway. 
First, we plan to demonstrate \system's effectiveness as an end-to-end framework on more complex downstream combination models like unsupervised LSCP \cite{zhao2019lscp} and supervised XGBOD \cite{zhao2018xgbod}. 
Second, we would further emphasize the interpretability provided by the pseudo-supervised approximation, which can be beyond feature importance provided in tree regressors. 
Third, there is room to investigate why and how the pseudo-supervised approximation could work in a more strict and theoretical way. 
% Specifically, we want to know how to choose supervised regressors and under what conditions the approximation could work. 
This study, as the first step, empirically shows that proximity-based models can benefit from the approximation.
, whereas linear models may not. 
% Fourth, other classical acceleration methods may be explored as well, e.g., focusing on numeric precision optimization \cite{mlsys2020_130}. 
Lastly, we may incorporate the emerging automated OD, e.g., MetaOD \cite{zhao2020automating}, to trim down the model space for further acceleration.
% Lastly, it is interesting to extend and generalize \system to other unsupervised tasks, such as streaming-based outlier detection and clustering ensemble. 
\clearpage
\newpage
\nocite{langley00}

\bibliography{ref}
\bibliographystyle{mlsys2021}

\clearpage
\newpage

%%%%%%%%%%%%%%%%%%%%%%%%%%%%%%%%%%%%%%%%%%%%%%%%%%%%%%%%%%%%%%%%%%%%%%%%%%%%%%%
%%%%%%%%%%%%%%%%%%%%%%%%%%%%%%%%%%%%%%%%%%%%%%%%%%%%%%%%%%%%%%%%%%%%%%%%%%%%%%%
% SUPPLEMENTAL CONTENT AS APPENDIX AFTER REFERENCES
%%%%%%%%%%%%%%%%%%%%%%%%%%%%%%%%%%%%%%%%%%%%%%%%%%%%%%%%%%%%%%%%%%%%%%%%%%%%%%%
%%%%%%%%%%%%%%%%%%%%%%%%%%%%%%%%%%%%%%%%%%%%%%%%%%%%%%%%%%%%%%%%%%%%%%%%%%%%%%%
\numberwithin{table}{section}
\appendix
\section*{Supplementary Material}
\textit{Details on Datasets and Models.}
\label{sec:appendix}

\section{Datasets and Setup}
\label{appendix:datasets}

Table \ref{table:data} describes the selected outlier detection benchmark datasets, and more than 20 outlier detection benchmark datasets are used in this study\footnote{ODDS Library: \url{http://odds.cs.stonybrook.edu}}\textsuperscript{,}\footnote{DAMI Datasets: \url{http://www.dbs.ifi.lmu.de/research/outlier-evaluation/DAMI}}. The data size $n$ varies from 452 (\textbf{Arrhythmia}) to 567,479 (\textbf{HTTP}) samples and the dimension $d$ ranges from 3 to 274. For both random projection and parallel scheduling experiments, the full datasets are used for model building (training). For the pseudo-supervised approximation experiments and the full framework assessment, 60\% of the data is used for training and the remaining 40\% is set aside for validation. For all experiments, performance is evaluated by taking the average of 10 independent trials using area under the receiver operating characteristic (ROC) curve and precision at rank $n$ (P@N)---here $n$ denotes the actual number of outliers. Both metrics are widely used in outlier research \cite{zimek2014ensembles,liu2019generative}.

\begin{table}[!htp]
\centering
	\caption{Selected real-world benchmark datasets} % title of Table
	\footnotesize
	\begin{tabular}{l | r  r  r  r } % centered columns (4 columns)
		\toprule %inserts double horizontal lines
		\textbf{Dataset} & \textbf{Pts (\textbf{\textit{n}})} & \textbf{Dim (\textbf{\textit{d}})} & \textbf{Outliers}	& \textbf{\% Outlier}\\
		\midrule
		Annthyroid & 7200  & 6   & 534   & 7.41 \\
		Arrhythmia & 452   & 274 & 66   & 14.60 \\
		Breastw    & 683   & 9   & 239  & 34.99 \\
		Cardio     & 1831  & 21  & 176  & 9.61  \\
		HTTP       & 567479	& 3	 & 2211 & 0.40  \\
		Letter     & 1600  & 32  & 100  & 6.25  \\
		MNIST      & 7603  & 100 & 700  & 9.21  \\
		Musk       & 3062  & 166 & 97   & 3.17  \\
		PageBlock & 5393  & 10  & 510  & 9.46  \\
		Pendigits  & 6870  & 16  & 156  & 2.27  \\
		Pima       & 768   & 8   & 268  & 34.90 \\
		Satellite  & 6435  & 36  & 2036 & 31.64 \\
		Satimage-2 & 5803  & 36  & 71   & 1.22  \\
		seismic    & 2584  & 10	 & 170  & 6.59 \\
		Shuttle    & 49097 & 9   & 3511 & 7.15  \\
		SpameSpace & 4207  & 57  & 1679 & 39.91 \\
        speech     & 3686  & 400 & 61	& 1.65 \\
% 		Stamps     & 340   & 9   & 31   & 9.12  \\
		Thyroid    & 3772  & 6   & 93   & 2.47  \\
		Vertebral  & 240   &  6  & 30   & 12.50 \\
		Vowels     & 1456  & 12  & 50   & 3.43  \\
        Waveform   & 3443  & 21  & 100  & 2.90  \\
% 		WBC        & 378   & 30  & 21   & 5.56  \\
		Wilt       & 4819  & 5   & 257  & 5.33  \\
		\bottomrule
	\end{tabular}
	\label{table:data} % is used to refer this table in the text
\end{table}

\section{Outlier Detection Models}
\label{appendix:model_set}

As shown in Table \ref{table:models_long}, we use a large group of outlier detection models in the experiment by varying algorithms and their corresponding hyperparameters. We build the model pool $\mathcal{M}$ by sampling from it.

\begin{table*}[!htp]
		\footnotesize
		\centering
		\caption{Outlier Detection Models in \system; see parameter definitions from PyOD \cite{zhao2019pyod}} % title of Table
	    \resizebox{\textwidth}{!}{% use resizebox with textwidth
			\begin{tabular}{lll} % centered columns (4 columns)
				\toprule
				\textbf{Method} &
				\textbf{Parameter 1} & \textbf{Parameter 2}\\
				\midrule
				ABOD \cite{kriegel2008angle}                           & n\_neighbors: $[3, 5, 10, 15, 20, 25, 50, 60, 70, 80, 90, 100]$   & N/A\\
				CBLOF \cite{he2003discovering} & n\_clusters: $[3, 5, 10, 15, 20]$ &  N/A\\
				Feature Bagging \cite{lazarevic2005feature}& n\_estimators: $[10, 20, 30, 40, 50, 75, 100, 150, 200]$ & N/A\\
				% COF \cite{tang2002enhancing}                           & n\_neighbors: $[3, 5, 10, 15, 20, 25, 50]$   & N/A \\
				HBOS \cite{goldstein2012histogram}                     & n\_histograms: $[5, 10, 20, 30, 40, 50, 75, 100]$  & tolerance: $[0.1, 0.2, 0.3, 0.4, 0.5]$ \\
				iForest \cite{liu2008isolation}	    	         	& n\_estimators: $[10, 20, 30, 40, 50, 75, 100, 150, 200]$  & max\_features: $[0.1, 0.2, 0.3, 0.4, 0.5, 0.6, 0.7, 0.8, 0.9]$ \\
				kNN \cite{Ramaswamy2000}                      & n\_neighbors: $[1, 5 ,10, 15, 20, 25, 50, 60, 70, 80, 90, 100]$   & method: ['largest', 'mean', 'median']\\
				% LODA \cite{pevny2016loda}                              & n\_bins: $[10, 20, 30, 40, 50, 75, 100, 150, 200]$        & n\_random\_cuts: $[5, 10, 15, 20, 25, 30]$\\
				LOF \cite{Breunig2000}                              & n\_neighbors: $[1, 5 ,10, 15, 20, 25, 50, 60, 70, 80, 90, 100]$   & method: ['manhattan', 'euclidean', 'minkowski']\\
				OCSVM \cite{scholkopf2001estimating}                   & nu (train error tol): $[0.1, 0.2, 0.3, 0.4, 0.5, 0.6, 0.7, 0.8, 0.9]$    & kernel: ['linear', 'poly', 'rbf', 'sigmoid']\\
				\bottomrule
				% 		\bottomrule
		\end{tabular}}
		\label{table:models_long} % is used to refer this table in the text
	\end{table*}
	
% \section{Additional Experiment Results}
% \label{appendix:psa_result}
% Here we present the additional comparison result of prediction performance between the original unsupervised models and pseudo-supervised approximators on 10 datasets with 6 costly algorithms by P@N (see \S \ref{EXP:PSA}). The approximators outperform in most cases.
% \input{tables/results_psa_prn}
%%%%%%%%%%%%%%%%%%%%%%%%%%%%%%%%%%%%%%%%%%%%%%%%%%%%%%%%%%%%%%%%%%%%%%%%%%%%%%%
%%%%%%%%%%%%%%%%%%%%%%%%%%%%%%%%%%%%%%%%%%%%%%%%%%%%%%%%%%%%%%%%%%%%%%%%%%%%%%%
% \clearpage
\vspace*{\fill}

\end{document}